\documentclass[11pt,a4paper]{article}

\usepackage[utf8]{inputenc}
\usepackage[T1]{fontenc}
\usepackage{mathptmx}      
\usepackage{microtype}       
\usepackage[margin=2.5cm]{geometry}
\usepackage{setspace}
\usepackage[english]{babel}
\usepackage{csquotes}

\usepackage{amsmath, amssymb}
\usepackage{booktabs}
\usepackage{array}
\usepackage{tabularx}
\newcolumntype{Y}{>{\raggedright\arraybackslash}X}
\newcolumntype{P}[1]{>{\raggedright\arraybackslash}p{#1}}
\usepackage{graphicx}
\usepackage{caption}
\usepackage{subcaption}
\usepackage{enumitem}         % For \begin{enumerate}[label=(\alph*)] etc.
\usepackage{placeins}

\usepackage[
    backend=biber,
    style=authoryear,
    maxcitenames=2,
    maxbibnames=99,
    giveninits=true,
    useprefix=true,
    uniquename=init,
    uniquelist=false,
    sorting=nyt,
    doi=true,
    isbn=false,
    url=true,
    dashed=false
]{biblatex}
\DeclareDelimFormat{nameyeardelim}{\addcomma\space}
\DeclareDelimFormat{nonameyeardelim}{\addcomma\space}

\usepackage{xcolor}
\usepackage[
    colorlinks=true,
    linkcolor=blue!50!black,
    citecolor=blue!50!black,
    urlcolor=blue!50!black,
    pdfstartview=FitH
]{hyperref}

\newcommand{\term}[1]{\emph{#1}}     
\newcommand{\limname}[1]{\textbf{#1}}  

\begin{document}

\title{%
    \textbf{Beyond Probabilistic Similarity:}\\[0.3em]
    \large Structural, Temporal, and Causal Limitations\\
    of Retrieval-Augmented Generation in the Legal Domain
}

\date{}
\author{Hudson de Martim\\ \small Federal Senate of Brazil, Brasília, Brazil\\ \small hudsonm@senado.leg.br}

\maketitle

% ---- Abstract ---------------------------------------------------------------
\begin{abstract}
Retrieval-Augmented Generation (RAG) has become a standard architectural response to unreliability in legal AI, yet high-profile failures, including fabricated citations submitted to courts and anachronistic legal content presented as current, continue to appear across jurisdictions. We argue that these failures are not residual confabulations to be eliminated by scaling language models, but symptoms of an architectural mismatch between probabilistic retrieval and the hierarchical, temporal, and institutional structure of legal knowledge. We develop the argument in three moves. First, we articulate the \textit{ontological commitment} of legal knowledge as a triad of properties derivable from classical legal theory: hierarchical and mereological structure, diachronic dynamism under operational closure, and causal traceability of institutional provenance grounded in the duty of justification. Second, we identify three corresponding pathologies of retrieval (\textit{mereological blindness}, \textit{diachronic blindness}, and \textit{causal opacity}), each developed with an operational definition, a failure mechanism, a canonical example, and detection criteria for diagnostic use. Third, we review the state of the art through this lens, showing that existing approaches address these requirements unevenly and do not yet compose into a paradigm that treats them as co-constitutive. From this analysis we derive four architectural commitments that characterize the \textit{deterministic-by-design} direction for legal retrieval: ontological primacy, event reification, bitemporal correctness, and deterministic interaction protocols. The framework concerns \textit{quaestio juris} (which norms apply and in what state) rather than the downstream tasks that act on identified norms, and addresses legislative and constitutional retrieval primarily, with interpretive time as an explicit extension.
\end{abstract}


\section{Introduction}
\label{sec:introduction}

In 2023, a federal judge in the Southern District of New York sanctioned two lawyers for submitting a brief in which some of the cited precedents did not exist. The citations had been generated by a chatbot, accepted without verification, and filed in court. The resulting U.S. proceeding became the canonical reference for a broader phenomenon~\parencite{mata2023avianca}. Similar incidents followed in the United Kingdom~\parencite{harber2023hmrc}, Canada~\parencite{zhang2024chen}, and Brazil~\parencite{tse2025multa}, all involving the same failure pattern: plausible-looking legal citations produced by AI systems, filed in legal proceedings, later found to be fabrications. Empirical studies confirmed that this is not an isolated operational failure but a systematic property of current legal AI: state-of-the-art language models confabulate legal content at measurable and legally consequential rates~\parencite{dahl2024largelegalfictions}, and subsequent evaluation of commercial retrieval-augmented legal research tools found that they continued to produce hallucinated content in a substantial fraction of queries despite retrieval grounding~\parencite{magesh2025hallucination}.

The response from the research community has been, for the most part, to address the symptom. A standard architectural intervention for improving grounding in language models is the Retrieval-Augmented Generation (RAG) paradigm introduced by \textcite{lewis2020retrieval}, which supplies the model at inference time with retrieved content from an external corpus. RAG reduces fabrication when the retriever returns relevant material, and legal applications of RAG are now the subject of a growing benchmark literature~\parencite{pipitone2024legalbench}. Yet the failures have not disappeared; we contend that they have changed shape. Where pre-RAG legal AI fabricated citations wholesale, RAG-augmented legal AI tends instead to cite real documents in ways that are anachronistic, structurally incomplete, or lacking the institutional grounding that legal fundamentals require---a pattern consistent with documented retrieval unreliability on large legal corpora~\parencite{reuter2025towards}. On this diagnosis, the problem migrates from the generator to the retriever.

This paper argues that the migration is not incidental. Dominant retrieval architectures --- vector search over chunked texts, and even Graph RAG variants that augment them with inferred entity--relation graphs~\parencite{edge2024graphrag,pan2024unifying} --- are calibrated for a paradigm that does not fit the legal domain. Graph RAG improves over flat retrieval by reintroducing structure, but the structure it reintroduces is inferred bottom-up from an LLM-derived entity--relation graph rather than inherited top-down from the formally decreed hierarchy of the legal instrument. More broadly, these architectures optimize for approximate semantic relevance: given a query, find the passages most similar to it in an embedding space. For open-domain corpora this is often adequate. For legal corpora it is not, because legal correctness is not a matter of semantic similarity. It is a matter of validity grounding: which norm was in force on a specific date, in a specific hierarchical context, by virtue of which institutional act. A retrieval function that cannot express these notions cannot guarantee legal adequacy, no matter how sophisticated its similarity metric or how large its training corpus.

The mismatch is not a gap to be closed by better embeddings or more retrieval stages. It is architectural. This paper's central contribution is to formulate the mismatch precisely and to derive from it both a diagnostic framework and a prescriptive direction. We do this in three moves.

First, we articulate what we call the \textit{ontological commitment} of legal knowledge (Section~\ref{sec:legal_domain}). Drawing on classical legal theory --- Kelsen's account of the legal order as a hierarchy of norms, Hart's distinction between object-level rules and the rules that govern their creation and change, and Luhmann's notion that a legal system changes only through its own recognized operations --- we identify three domain-level properties that any computational treatment of legal content must respect: hierarchical and mereological structure, diachronic dynamism under operational closure, and causal traceability, understood as traceability of institutional provenance grounded in the duty of justification. These are not stipulative definitions. They are conditions of legal adequacy derived from the theoretical apparatus through which the legal order understands itself, reformulated as requirements on retrieval architectures.

Second, we name three pathologies of retrieval corresponding to systematic failures of these commitments (Section~\ref{sec:three_limitations}). \textit{Mereological blindness} is the failure to preserve the part--whole structure of legal texts. \textit{Diachronic blindness} is the failure to reconstruct the state of norms at specific points in time, encompassing both statutory evolution through amendments and interpretive evolution through binding decisions. \textit{Causal opacity} is the failure to expose, for each returned provision or material synthesized claim, the institutional chain of acts that grounds its validity. The pathologies are architecture-agnostic: they characterize what retrieval output lacks, not which technology produced it. We provide for each an operational definition, a failure mechanism, a canonical example, and detection criteria that allow practitioners to diagnose the pathology in existing systems.

Third, we review the state of the art --- including the classical AI \& Law traditions of rule representation, case-based reasoning, and legal information retrieval --- through the lens of the pathologies rather than through the lens of technologies (Sections~\ref{sec:review} and~\ref{sec:toward_neurosymbolic}). The review shows that the pathologies are addressed unevenly by identifiable lines of work: document standards, structure-aware chunking, graph and ontology RAG, temporal knowledge graphs, event-centric modeling, provenance modeling, explainability, and tool-using agents. These partial solutions, however, do not compose automatically. Their architectural assumptions differ, and retrofitting any two requirements onto a substrate not designed for co-satisfaction can mask or amplify the remaining pathology. Joint satisfaction requires a different architectural posture, which we formalize through four commitments: ontological primacy, event reification, bitemporal correctness, and deterministic interaction protocols.

Together, these commitments characterize what we call \term{deterministic-by-design} legal retrieval, an architectural direction better suited to the subset of legal applications in which validity grounding, temporal correctness, and institutional provenance are load-bearing requirements. The term does not mean that every component of the system must be deterministic, nor that language models are excluded. It means that legality-critical retrieval operations --- structural traversal, point-in-time resolution, and provenance-chain reconstruction --- are governed by explicit, reproducible, and auditable mechanisms rather than by unconstrained probabilistic inference. Language models may still be used for natural-language understanding, planning, and synthesis, provided that validity grounding is supplied by the substrate rather than inferred by the model.

\paragraph{Methodological note.}
This paper is a critical and theory-driven survey rather than a systematic mapping review. The reviewed literature was selected by functional relevance to the three domain-level commitments developed in Section~\ref{sec:legal_domain}: preservation of legal structure, reconstruction of temporal state, and exposure of institutional provenance. We include work from RAG, Graph RAG, legal document standards, legal ontologies, temporal knowledge graphs, provenance modeling, explainable AI, and tool-using agents insofar as each strand addresses one of the diagnostic pathologies. The aim is not exhaustive coverage of legal AI, but a pathology-oriented synthesis of the architectural assumptions that determine legal adequacy at the retrieval layer. The paper does not report a benchmarked empirical comparison of existing systems; its contribution is formal and diagnostic, specifying criteria intended to support such comparisons in subsequent work.

\paragraph{Scope and delimitations.}
Two delimitations bound the analysis. First, the framework concerns \textit{quaestio juris}, the question of which norms apply and in what state. It does not extend to other dimensions of legal practice in which AI systems face structural limitations, including \textit{quaestio facti}, the question of what happened (whether real-world events occurred as claimed, whether witnesses are credible, whether evidence establishes a state of affairs), as well as judicial discretion, evaluative judgment on open-textured concepts, and other tasks in which legal practice depends on contextual or moral reasoning rather than on the identification of applicable norms. Recent analyses of generative AI in legal reasoning have argued that its limits across several of these dimensions may be structural rather than technical~\parencite{linna2026challenges}. Our framework concerns the prior question, in the sense that any of these subsequent operations presupposes the correct identification of available norms: what norms are available, in what version, and with what provenance, over which factual disputes are then adjudicated and judgment is exercised.

Second, our primary focus is on legislative and constitutional texts. This is a methodological delimitation, not a claim that case law is secondary. In many jurisdictions, judicial decisions are themselves primary sources of law with distinctive structural and temporal properties. Within the scope of this paper, we argue that the three pathologies and the four architectural commitments apply to case law as well, and the interpretive-time dimension of diachronic blindness (Sections~\ref{subsec:diachronic} and~\ref{subsec:addressing_diachronic}) explicitly engages the temporal structure of binding judicial interpretation. Full treatment of case law as a first-class domain is identified as a research direction in Section~\ref{sec:agenda}.

\paragraph{Contributions.}
The paper makes four contributions:

\begin{enumerate}[leftmargin=2em, itemsep=0.5ex]
    \item \textbf{Ontological commitment of legal knowledge.} A formulation of three domain-level properties --- hierarchical and mereological structure, diachronic dynamism under operational closure, and causal traceability of institutional provenance --- derivable from classical legal theory and translatable into retrieval requirements.

    \item \textbf{Diagnostic framework of three pathologies.} The identification and formal development of \textit{mereological blindness}, \textit{diachronic blindness}, and \textit{causal opacity} as reusable pathologies of retrieval, each with operational definitions, failure mechanisms, canonical examples, and detection criteria.

    \item \textbf{Pathology-organized critical review.} A review of current legal retrieval approaches organized by which pathology each addresses, with an explicit argument that partial solutions do not compose automatically when retrofitted onto substrates designed for other purposes.

    \item \textbf{Architectural commitments.} The specification of \textit{ontological primacy}, \textit{event reification}, \textit{bitemporal correctness}, and \textit{deterministic interaction protocols} as mutually reinforcing commitments characterizing deterministic-by-design legal retrieval, together with an account of their costs and limits.
\end{enumerate}

\paragraph{Roadmap.}
Section~\ref{sec:legal_domain} establishes the ontological commitment of legal knowledge. Section~\ref{sec:three_limitations} develops the three-pathology framework. Section~\ref{sec:review} reviews current approaches through that framework. Section~\ref{sec:toward_neurosymbolic} derives the four architectural commitments, discusses existing instances, and addresses trade-offs. Section~\ref{sec:agenda} identifies four research directions. Section~\ref{sec:conclusion} concludes.

% =============================================================================
\section{The Legal Domain as an Ontological Challenge}
\label{sec:legal_domain}

Before diagnosing the failures of generic information-retrieval architectures in legal settings, it is necessary to articulate what makes law a distinctive object of computational treatment. Legal texts are not narrative prose, encyclopedic description, or scientific documentation. They constitute a structured normative system whose validity conditions are neither semantic nor probabilistic, but formal and institutional. This section identifies three domain-level properties of legal systems that any retrieval architecture must respect if it is to avoid returning legally inadequate content: (i)~hierarchical and mereological structure; (ii)~diachronic dynamism under operational closure; and (iii)~causal traceability of institutional provenance, grounded in the duty of justification. These properties are drawn from classical legal theory and reformulated as computational requirements. Together they constitute what we term the \term{ontological commitment} of legal knowledge: a set of requirements that subsequent sections will use as benchmarks to diagnose specific architectural failures.

\subsection{Hierarchical Normativity and Mereological Structure}
\label{subsec:hierarchy}

The modern positivist tradition, following Kelsen's formulation of the legal order as a stepped structure (\textit{Stufenbau}, a layered hierarchy of norms)~\parencite{kelsen1967pure}, conceives legal validity as a cascading relation: each norm derives its validity from a higher one, terminating in a presupposed fundamental norm. The practical consequence for information retrieval is not that the retriever should adjudicate validity itself, but that it must mobilize validity status the legal order has already established. The retrieval substrate must therefore represent hierarchical relations, enabling authorities, and authoritative validity-affecting events, including invalidations, suspensions, repeals, and recognized conflicts. Where such status has been recorded, retrieval should expose it together with the affected norm rather than return the text in isolation; a system that ignores represented status may present textually current content as applicable even though it has been limited, suspended, or invalidated by an authoritative act. Where the substrate records that an incompatibility between norms has been raised but not authoritatively resolved, retrieval should expose that fact so that downstream interpretation can address it. The detection of such conflicts is itself a matter of legal-domain curation or authoritative recognition, not of inference by the retrieval function.

This hierarchical dimension, which we shall call \term{systemic hierarchy}, coexists with a second dimension whose retrieval implications are equally consequential: \term{mereological hierarchy}. Legal texts are organized through strict part--whole relations governed by formalized drafting techniques~\parencite{Palmirani2011,brazil1998lawdrafting}. A statute is partitioned into books, titles, chapters, sections, articles, paragraphs, items, and sub-items, each level subordinating the next. The meaning of any given provision depends on its position within this tree. A provision read in isolation from its \textit{caput}, the opening or governing clause that heads an article and sets the conditions under which its dependent items operate, is syntactically incomplete and semantically ambiguous.

A concrete example illustrates the point. Item~VI(b) of Article~150 of the Brazilian Federal Constitution, in the wording established by Constitutional Amendment~132/2023~\parencite{brazil2023ec132}, refers to ``religious entities and temples of any cult, including their assistance and charitable organizations.'' Read in isolation, this phrase names a class of entities but asserts no legal consequence. Its normative content emerges only from the governing \textit{caput} and intermediate items, which identify the public entities bound by the prohibition and the act they are prohibited from performing. The fragment becomes legally operative only within that mereological context.

This dependency is not a drafting quirk; it is a structural feature of legal technique~\parencite{simons1987parts}. We can describe the corresponding retrieval failure as \term{decontextualized normativity}: a legal fragment outside its part--whole hierarchy is not merely incomplete; it is potentially misleading, suggesting rules that do not exist or obscuring conditions that do.

Hart's distinction between primary and secondary rules~\parencite{Hart2012} further refines this picture. Secondary rules, or rules about rules, include \term{rules of change}, which specify how norms are created, modified, or repealed, and \term{rules of recognition}, which specify the criteria of validity within the system. A computational representation of law that ignores this meta-structure cannot distinguish a norm that is valid from one that merely happens to be prevalent in a corpus. The corpus is evidence; the rule of recognition is what confers validity.

\subsection{Diachronic Dynamism and Operational Closure}
\label{subsec:diachronic}

The second defining property of legal systems is diachronic dynamism under operational closure. Law is not a static repository; it is continuously produced and modified. But not any modification counts. The notion of autopoiesis --- a system that reproduces and modifies itself only through its own internal operations --- introduced by Maturana and Varela~\parencite{Maturana1980} and adapted to legal sociology by Luhmann~\parencite{Luhmann2004} captures this point precisely: the legal system operates under \term{operational closure}. Its elements can be modified only by operations internal to the system: legislative acts, constitutional reforms, binding judicial decisions, or other formally recognized legal operations. Social pressure, political preference, or statistical regularity in textual usage do not alter legal validity. They must be translated into the system's syntax through a legally recognized act to produce structural effect.\footnote{We use operational closure only as an abstract modeling constraint: admissible validity transitions are transitions triggered or recognized by events belonging to a formally defined class of legal operations. Nothing in the argument depends on accepting Luhmann's broader sociology of law.}

This has a sharp computational consequence. A retrieval system cannot infer the obsolescence of a norm from topical trends, news mentions, or proximity in a vector space. Validity transitions are event-driven: a norm is in force at time $t$ only if a legally recognized event at time $t' \leq t$ created, activated, or fixed the relevant state, and no subsequent legally recognized event has closed, suspended, superseded, or otherwise altered that state. Anything else is an inference about evidence, not about law.

The temporal architecture of legal systems admits further refinement. Each norm has a lifecycle bounded by formal events: enactment, publication, \textit{vacatio legis} (a deferral period between publication and legal effect), the validity interval itself, and termination. Termination may result from express repeal, tacit repeal, expiration of a temporary statute, suspension, declaration of invalidity, or authoritative interpretation. Tacit repeal and tacit overruling complicate the event-driven picture because the validity-affecting transition may be recognized through doctrinal inference rather than declared in a single explicit act. In such cases, deterministic retrieval should not infer the transition silently: it should either rely on an authoritative recognition event or flag the transition as requiring human/legal interpretation. A unified model of legal time must represent these causes without collapsing their differences.

A further complication is specific to legislative technique: amendments operate by \term{difference}, not by restatement~\parencite{palmirani2011legislative}. An amending statute typically states only the new wording, without transcribing the replaced text. To reconstruct the state of a norm at a past date, a system must therefore chain a sequence of differential operations, identifying which amendments affected which provisions at which dates. Each step is a formal operation over a history of legal events, not a heuristic comparison over textually similar passages.

Common-law jurisdictions and constitutional adjudication add another dimension to this dynamism. The text of a provision may remain fixed while its authoritative interpretation shifts, what civil-law systems often describe as constitutional mutation (a change in operative meaning without a change in text) and what common-law systems experience through overruling, distinguishing, or changes in controlling authority. The literal text of a constitutional clause may remain stable while its operative meaning changes through binding decisions. A retrieval system that targets only statutory text, whether as a current consolidation or as a sequence of textual versions, captures only part of the normative state of the system. The prevailing interpretation, when it differs from the literal text, is itself part of what counts as ``the law'' at time $t$.

The consequence is that the temporal dimension of legal knowledge is \term{bitemporal} in a strict technical sense~\parencite{snodgrass1999developing}: it requires distinguishing \term{valid time}, when a norm produces legal effect, from \term{transaction time}, when a norm was recorded in the official system. For legal corpora, valid time itself often decomposes further: legal informatics standards such as the European Legislation Identifier~\parencite{eli_ontology} represent date of entry into force and date of applicability as distinct properties, since a norm may enter the legal order on one date while producing its effects on another, as in \textit{vacatio legis}, deferred effectiveness, or retroactive application. The bitemporal framework accommodates these sub-distinctions as additional properties on the valid-time interval; what matters for retrieval is that all relevant temporal dimensions are independently representable and independently queryable, so that the system can answer not only ``what was the text on date $T$?'' but also ``was that text in force on $T$?'' and ``was it applicable to events on $T$?''.

In legal retrieval, this distinction must also accommodate the possibility that statutory time and interpretive time diverge. Architecturally, interpretive time need not introduce a wholly separate temporal axis: a binding judicial decision can be modeled as a validity-affecting event that alters the operative interpretive state of a provision while the textual state remains unchanged. Collapsing these dimensions produces anachronism: the citation of a text that was, technically, on the books, but was not the operative norm on the date in question.

\subsection{Causality and the Duty of Justification}
\label{subsec:causality}

The third domain-level property, decisive for legal applications, is the duty of justification. Legal decisions must state their reasons. In the Brazilian constitutional order, Article~93, item~IX requires judicial decisions to be reasoned under penalty of nullity; analogous obligations are found in virtually every jurisdiction operating under rule-of-law commitments. This is not a procedural nicety. It expresses a deeper commitment: the legitimacy of a legal outcome depends on the reconstructibility of the reasoning and authority that produced it~\parencite{schauer2009thinking}.

Applied to retrieval, this duty translates into a demand for causal traceability of institutional provenance. A terminological note is in order: throughout this paper, \term{causal} refers to institutional provenance, the chain of official acts that produce and transform the validity or operative state of a norm, and not to the statistical notion of causal inference over probabilistic graphical models familiar from contemporary machine learning~\parencite{pearl2009causality}. The two notions share a word but not a referent. Institutional causality is documentary and authority-based: a specific act creates, modifies, interprets, suspends, or terminates a specific legal state at a specific time.

Returning a norm is therefore not enough. A legally usable system must state why this norm, in this version or operative interpretation, applies to this date. Each such justification invokes a chain: the norm was enacted by event $E_1$, modified by event $E_2$, and, where applicable, interpreted, suspended, or terminated by event $E_3$. Each event is traceable to a specific official act, with a date, a competent authority, a legal effect, and a recorded source. The retrieval system must either reproduce this chain or fail transparently, acknowledging what it cannot ground.

This requirement exceeds what explainability frameworks typically demand of AI systems~\parencite{richmond2024explainable}. Post-hoc explanation of model behaviour is insufficient when the system is supposed to produce legally relevant outputs. The relevant question is not only ``why did the system produce this answer?'' but ``what chain of institutional acts gives this answer legal force?'' A retrieval architecture that returns the correct text but cannot name the events that instituted, modified, or authorized it has failed not merely at explanation, but at grounding.

\subsection{Implications for Computational Treatment}
\label{subsec:implications}

The three properties just outlined, hierarchical and mereological structure, diachronic dynamism under operational closure, and causal traceability of institutional provenance, jointly define the ontological commitment of legal knowledge. Any computational system that claims to retrieve, reason about, or generate legal content must either satisfy these commitments or acknowledge openly that it does not.

This formulation is not a matter of disciplinary preference. It is a condition of legal adequacy. A system that returns current text while ignoring mereological context returns decontextualized normativity. A system that returns the literal text of a provision without reconstructing its temporal state returns anachronism. A system that returns the correct provision without naming the acts that ground its validity or operative interpretation returns ungrounded legal information. In any of these failure modes, the output may appear fluent, relevant, and textually correct. It nonetheless fails as legal information.

Three clarifications are in order. First, the commitments above concern validity grounding, not factual truth. The framework addresses \textit{quaestio juris}: what norms, versions, interpretations, and authorities are available for legal reasoning. It does not address \textit{quaestio facti} --- whether a given real-world event occurred, whether a witness is credible, whether a defendant was in a given place at a given time --- nor does it extend to judicial discretion, evaluative judgment on open-textured concepts, or other dimensions of legal practice that operate downstream of norm identification. A deterministic legal retrieval substrate provides the normative material over which these subsequent operations are exercised; it does not perform them.

Second, the commitments do not imply that every legal question has a mechanically determinate answer. Legal interpretation remains interpretive, and disagreement remains possible. The point is narrower: where the legal system records formal acts, temporal states, structural relations, and authoritative sources, a retrieval system should preserve and expose them rather than replace them with approximate semantic relevance. Determinism here means reproducibility and auditability of retrieval operations over a represented substrate, not the elimination of legal judgment.

Third, the jurisprudential scope of the argument is deliberately bounded. The three domain-level commitments are most naturally compatible with positivist or institutionally oriented accounts of law: accounts that treat legal validity as traceable to formal criteria of authority, hierarchy, procedure, and temporal effect, in the tradition represented by Kelsen, Hart, and related strands. They are less directly aligned with strongly realist accounts that locate legal meaning primarily in patterns of judicial behaviour, or with critical accounts that emphasize how codified rules are shaped by deeper political, social, or distributive dynamics. We do not argue against those accounts here. Rather, we delimit the target of the framework. The commitments developed in this paper are designed for retrieval over legal materials as they are formally codified, versioned, and institutionally maintained. That scope covers statutory and constitutional corpora in many contemporary jurisdictions, as well as the substantial fraction of legal practice that depends on identifying applicable texts, versions, authorities, and validity periods within those corpora. A retrieval architecture adequate to that scope is not thereby a general theory of law; it is an engineering response to a specific and tractable subset of legal information needs.

The remainder of this paper takes these commitments as benchmarks. We argue that the dominant information-retrieval paradigm, including the way Retrieval-Augmented Generation is typically deployed in legal-tech products, does not satisfy these commitments as a default architectural property. Specific systems may go further, but the failure is not incidental: it is traceable to choices built into the paradigm itself. Section~\ref{sec:three_limitations} formalizes the three failure modes as a diagnostic framework.

% =============================================================================
\section{The Three Limitations: A Diagnostic Framework}
\label{sec:three_limitations}

Section~\ref{sec:legal_domain} defined three commitments that legal information must satisfy: preservation of hierarchical and mereological structure, reconstruction of legal state across time under operational closure, and exposure of the institutional event chain that grounds validity claims. This section inverts the perspective. We ask how retrieval systems fail to satisfy these commitments. The answer is not a single failure mode but three: \limname{mereological blindness}, \limname{diachronic blindness}, and \limname{causal opacity}.

We treat these as \term{pathologies of retrieval}: properties of a system's output that render it inadequate as legal information, regardless of how fluent, topical, or textually accurate the output appears. The pathologies are architecture-agnostic. Vector-based retrieval, graph-based retrieval, keyword-indexed systems, and purely generative systems can each exhibit any of the three. Each subsection below provides an operational definition, a failure mechanism, a canonical example, and detection criteria. Sections~\ref{sec:review} and~\ref{sec:toward_neurosymbolic} then connect the pathologies to architectural choices.

\subsection{Mereological Blindness}
\label{subsec:mereological_blindness}

\paragraph{Definition.}
Mereological blindness is the failure of a retrieval system to preserve, recover, or respect the part--whole structure of legal texts in its outputs. A system exhibits mereological blindness when the fragments it returns do not carry enough of their governing context for their legal meaning to be reconstructed, or when queries scoped to a hierarchical unit cannot be resolved with guaranteed completeness within that unit.

\paragraph{Failure mechanism.}
The mechanism has two variants, which often co-occur. The first is \term{representational}: the system encodes legal texts at a granularity that does not match legal drafting. Fixed-size chunking, paragraph-level segmentation that ignores legal-dispositive boundaries, and sub-article chunking that isolates items or sub-items from their governing \textit{caput} all instantiate this variant. The result is a corpus in which legally dependent fragments are treated as semantically independent units. The second variant is \term{operational}: even when structure is represented in the index, the retrieval function does not enforce ancestor closure. A system may record that item~VI(b) of Article~150 is subordinate to the article's \textit{caput} and yet return or pass to the generator only the sub-item. The structural relations exist in the data model but are not mobilized by retrieval, reranking, or context assembly.

Both variants share a common root: the retrieval function is optimized for relevance, and relevance is computed as a property of fragments rather than as a property of fragments-in-context. Standard vector similarity exemplifies the mismatch. It is symmetric over passage pairs, whereas legal mereology is directional: a sub-item depends on its governing \textit{caput} to acquire normative meaning, whereas no particular sub-item governs the \textit{caput} in the same way. Asymmetric retrievers and cross-encoders can score query--passage pairs directionally, but they do not by themselves encode the legal direction of dependence among provisions. Similarity can register proximity; it cannot, without structural relations or deterministic expansion, enforce ancestor closure. The pathology is therefore not poor tuning. It is that similarity-based tooling treats as reciprocal a relation that legal drafting makes directional.

\paragraph{Canonical example.}
Consider a user querying: \textit{``Does Brazilian law exempt religious organizations from taxes?''} A retrieval system with mereological blindness may return the following fragment from Article~150 of the Brazilian Federal Constitution:

\begin{quote}
\textit{``b) religious entities and temples of any cult, including their assistance and charitable organizations.''}
\end{quote}

\noindent Returned in isolation, the fragment names a class of entities but specifies neither the legal consequence nor the public entities bound by the prohibition. Its meaning depends on the governing chain: Article~150's \textit{caput} identifies the addressees of the prohibition, and item~VI identifies the prohibited act as levying taxes on the enumerated classes. Only within that chain does ``religious entities and temples of any cult'' acquire the normative status of being protected from taxation. A system that returns the sub-item without its ancestors has produced text that is authentic but legally underdetermined.

The pathology also appears at coarser granularities. An article-level chunk avoids the item-in-isolation problem but may lose dependencies that cross articles, such as definitions, exceptions, and cross-references. In such cases, the pathology re-emerges at a different scale. Avoiding it requires retrieval to expand context on demand through typed legal relations, not to commit to any fixed level of granularity.

\paragraph{Detection criteria.}
A system can be tested for mereological blindness along three axes. These criteria are substrate-facing but retrieval-oriented: a language model or agent may invoke the relevant operations, but satisfaction depends on typed relations and deterministic retrieval behaviour, not on unconstrained inference by the agent.

\begin{enumerate}[label=(\alph*), leftmargin=2em, itemsep=0.5ex]
    \item \textbf{Descendant completeness under scope queries.} When a query is formulated as ``all items of $X$'' or ``all sub-items of $Y$'', does the system guarantee an exhaustive set within the represented hierarchy, returned in canonical structural order, or does it approximate via a Top-$K$ cut that may omit relevant children?

    \item \textbf{Ancestor closure and subordination signaling.} Given a retrieved fragment, can the system recover and, when the fragment is used as legal support, include or attach the governing elements under which it falls, such as the \textit{caput} (the governing clause heading an article), containing article, section, or chapter? When those governing elements are not included, does the system expose that the fragment is subordinate within the represented hierarchy rather than presenting it as contextually complete?

    \item \textbf{Boundary and expansion integrity.} For queries whose scope is restricted to a hierarchical unit, are returned fragments either strictly within that unit or explicitly justified by typed legal relations represented in the substrate, rather than leaking across boundaries on the basis of semantic similarity alone?
\end{enumerate}

Satisfying all three criteria addresses mereological blindness with respect to the represented hierarchy; failing any one exhibits the pathology to some degree.

\subsection{Diachronic Blindness}
\label{subsec:diachronic_blindness}

\paragraph{Definition.}
Diachronic blindness is the failure of a retrieval system to reconstruct the state of normative content at a specific point in time, or to distinguish among coexisting versions of the same provision whose differences matter for legal purposes. A system exhibits diachronic blindness when its output is temporally unqualified, temporally ambiguous, or temporally anachronistic with respect to the query it is answering.

\paragraph{Failure mechanism.}
The mechanism is representational at its root: the system treats legal text as a static object rather than as the current state of an evolving process. Five modes instantiate this. First, the corpus may contain only current versions, so past-date queries default to anachronism. Second, the corpus may preserve original instruments and amendments but not the consolidated structural-unit states that result from applying them; the information exists, but reconstruction is delegated to probabilistic reasoning at query time. Third, the corpus may be versioned only at document granularity, producing whole-document snapshots that obscure structural-unit differentials and generate near-duplicate retrieval results unless temporal filtering is enforced. Fourth, even structural-unit versions may be untethered from the events that produced them, preventing event-bounded queries such as ``before the amendment introduced by Act~Y.'' Fifth, a system that handles statutory time may still miss \term{interpretive time}: binding decisions can alter the operative norm without altering the text.

Across these modes, the underlying cause is the same: the system lacks a formal model in which time is a first-class dimension of retrieval and, for full legal adequacy, legal events are the operators that transition one state to another~\parencite{snodgrass1999developing}.

\paragraph{Canonical example.}
Article~6 of the Brazilian Federal Constitution enumerates social rights. Enacted in 1988 with an initial list of rights, it has been amended several times: housing was added by Constitutional Amendment~26/2000, food by EC~64/2010, transportation by EC~90/2015, and a basic-income provision by EC~114/2021~\parencite{brazil1988constitution,brazil2000ec26,brazil2010ec64,brazil2015ec90,brazil2021ec114}.

Consider a user asking: \textit{``Did the right to housing exist in Brazilian constitutional law in 1999?''} The correct answer is negative: housing was added in 2000. A diachronically blind system may return the current text of Article~6, which includes housing; return multiple versions without temporal grounding; or delegate the reconstruction of the 1999 state to probabilistic reasoning over authentic but separately retrieved fragments. Each outcome may look plausible, but each fails the target-date query.

The example also illustrates why similarity is insufficient. Successive versions of Article~6 share most of their tokens and legal vocabulary; their embeddings are therefore close in latent space. Without temporal filtering or event-bounded versioning, a similarity-based retriever has no principled way to select the version operative at the queried date.

Diachronic blindness also extends to interpretive time. Article~5, item~LVII of the Brazilian Constitution has been textually stable since 1988, but its operative meaning on imprisonment after second-instance conviction shifted repeatedly through binding decisions, from HC~68.726 in 1991 to the ADCs~43, 44, and 54 in 2019, none of which altered the text. A system that retrieves the unchanged text and stops there is temporally blind to the interpretive events that shaped the operative state.

\paragraph{Detection criteria.}
A system can be tested for diachronic blindness along five axes:

\begin{enumerate}[label=(\alph*), leftmargin=2em, itemsep=0.5ex]
    \item \textbf{Point-in-time recovery.} Given a provision and a query date $T$ within the historical interval covered by the corpus, can the system return the text that was in force at $T$, or state that no version was in force at that date, rather than defaulting to the current version, returning multiple temporally indistinguishable versions, or delegating the temporal selection to a probabilistic reconstruction step?

    \item \textbf{Bitemporal correctness.} Does the system distinguish the date on which a norm was \textit{recorded} in the official system (transaction time) from the date on which it \textit{is in force} (valid time)? This distinction matters precisely for edge cases in which the two times diverge: \textit{vacatio legis} periods, retroactive provisions, delayed effectiveness, and corrigenda to the official record.

    \item \textbf{Materialized structural-unit versioning.} Are structural-unit states materialized, or deterministically materializable from explicit transformation events, rather than reconstructed probabilistically from original instruments and amendment texts at query time?

    \item \textbf{Event-bounded validity.} For each returned textual version, can the system identify the legal event that opened its validity interval and, where applicable, the event that closed, superseded, or suspended it? This criterion concerns temporal grounding of version intervals, not the full provenance-chain reconstruction addressed under causal opacity.

    \item \textbf{Interpretive--statutory coherence.} When binding or otherwise authoritative interpretation has shifted without textual amendment, does the system surface both the literal text and the operative interpretation for the queried date, or flag that the operative state depends on interpretive authority rather than textual change alone?
\end{enumerate}

Criterion~(a) captures primitive point-in-time awareness; criteria~(a)--(d) capture mature legislative or constitutional textual time; all five are required for diachronic adequacy where interpretive time is within scope.

\subsection{Causal Opacity}
\label{subsec:causal_opacity}

\paragraph{Definition.}
Causal opacity is the failure of a retrieval system to expose, for each returned provision, operative interpretation, or material synthesized claim, the chain of institutional events that grounds it. A system exhibits causal opacity when its outputs cannot be audited back to specific, named acts --- statutes, amendments, rulings, administrative acts --- that gave the returned content its legal force.

\paragraph{Failure mechanism.}
Causal opacity arises from an architectural mismatch: legal retrieval must support justified decisions, but retrieval systems are typically optimized for relevance. Relevance, whether lexical, semantic, topical, or learned, is a property of the match between a query and a document. It is not an answer to the question: \textit{what institutional act grounds this content?}

The mismatch has four expressions. First, probabilistic retrievers return similarity scores, not provenance chains. Second, generative synthesis dissolves boundaries among retrieved fragments, so even fragment-level provenance is not preserved by default at the level required for legal justification; claim-level or span-level attribution must be designed explicitly. Third, post-hoc explanation techniques explain model behaviour, not legal grounding~\parencite{richmond2024explainable}. They can show why the model produced an output, but not which amendment, ruling, or administrative act grounds the legal state asserted by the output. Fourth, metadata can remain non-operative. Amendment identifiers, dates, and jurisdictional markers matter, but they do not by themselves produce an auditable chain unless retrieval composes and surfaces them as part of the returned state, attached audit artifact, or stable follow-up operation.

Causal transparency is therefore a property of what the system exposes for audit, not merely of what it stores, retrieves, or internally computes. The criterion is reconstructibility: can the user follow the sequence of institutional acts from enactment, through amendments and interpretive reversals, to the queried state?

\paragraph{Canonical example.}
Return to Article~5, item~LVII of the Brazilian Federal Constitution, the presumption of innocence. A user asks: \textit{``What is the current binding interpretation of the presumption of innocence under Brazilian constitutional law?''}

A causally opaque system may return the literal text, perhaps accompanied by related precedents ranked by similarity. The user receives content, but not grounding: which ruling is authoritative, which has been superseded, and which sequence of decisions fixes the operative interpretation.

A causally transparent system returns a chain: the original text as enacted in 1988; HC~68.726 in 1991, holding that the presumption of innocence did not bar execution after a second-instance conviction; its reversal by HC~84.078 in 2009, which required final judgment before execution; the counter-reversal by HC~126.292 in 2016; and the restoration of the 2009 position by ADCs~43, 44, and 54 in 2019\footnote{A later development illustrates the same point. In 2024, the Federal Supreme Court held (Tema~1.068, RE~1.235.340) that the sovereignty of jury verdicts permits immediate execution of jury-imposed sentences, creating a context-specific exception grounded in a different constitutional clause (Art.~5, XXXVIII) rather than reversing the general rule restored by the 2019 declaratory actions. Answering ``what is the operative interpretation?'' therefore now requires distinguishing the procedural context --- precisely the kind of contextual complexity a deterministic substrate must represent rather than collapse.}. Each step is named, dated, attributable to a specific authority, and auditable against primary sources. The example illustrates that causal opacity is not binary: a system may surface the most recent event but not the full chain, legislative events but not interpretive ones, or dates and authorities without the legal effect each event introduced.

\paragraph{Detection criteria.}
A system can be tested for causal opacity along five axes. These criteria are substrate-facing: they test whether institutional provenance is represented in the substrate as a structural property and made available by retrieval, rather than reconstructed by downstream components through inference over text.

\begin{enumerate}[label=(\alph*), leftmargin=2em, itemsep=0.5ex]
    \item \textbf{Grounding-event attribution.} Given a returned provision or operative interpretation, can the system identify the statute, amendment, ruling, administrative act, or other authoritative event that produced, modified, or fixed the returned state?
    
    \item \textbf{Full lineage reconstruction.} For a provision with at least two amendments or interpretive shifts, can the system return the ordered sequence from original enactment through each validity-affecting event to the queried state, including events that create, amend, repeal, suspend, supersede, or authoritatively interpret the provision?
    
    \item \textbf{Dual-chain surfacing.} When the substrate represents both textual versions of a provision and interpretive events that have altered its operative state, does the retrieval protocol expose both chains together --- the statutory version chain and the judicial or administrative authority chain --- so that downstream components can identify whether they diverge for the queried date?
    
    \item \textbf{Claim-level provenance traceability.} Does the substrate represent provenance at a granularity that permits each material legal claim, when composed by downstream components, to be traceable to its specific supporting provision, version, interpretation, and grounding event? This requires that provisions, versions, interpretations, and events are individually addressable and that the relations between them are explicit.
    
    \item \textbf{Auditable chain composition.} Are the elements of a provenance chain --- source identifiers, dates, authorities, event types, and the relations linking them --- represented in the substrate as identifiable and stable entities, so that a chain can be deterministically composed and independently checked against primary sources?
\end{enumerate}

A system failing even criterion~(a) is causally opaque at the most basic level. A system passing criteria~(a)--(c) supports retrieval-side traceability of institutional provenance. A system passing all five additionally provides substrate-level conditions for claim-level traceability and deterministic audit-chain composition by downstream components.

\subsection{The Framework in Summary}
\label{subsec:framework_summary}

The three pathologies map primarily onto the three domain-level commitments established in Section~\ref{sec:legal_domain}: mereological blindness violates the commitment to hierarchical and mereological structure; diachronic blindness violates the commitment to operational closure and temporal coherence; causal opacity violates the commitment to causal traceability of institutional provenance. The mapping is not accidental; it is the point of the framework. Table~\ref{tab:pathologies} summarizes the diagnostic structure.

\begin{table}[!ht]
\centering
\caption{Pathologies, diagnostic tests, and architectural responses.}
\label{tab:pathologies}
\footnotesize
\begin{tabularx}{\linewidth}{P{0.16\linewidth}P{0.18\linewidth}Y Y P{0.18\linewidth}}
\toprule
\textbf{Pathology} & \textbf{Violated commitment} & \textbf{Typical failure} & \textbf{Diagnostic test} & \textbf{Architectural response} \\
\midrule
Mereological blindness & Hierarchical and mereological structure & Returns a dependent fragment without its governing context & Ancestor closure, descendant completeness, and boundary integrity & Ontological primacy and typed structural traversal \\
Diachronic blindness & Operational closure and temporal coherence & Returns a version that is anachronistic, ambiguous, or inferred without valid-time confirmation for a target-date query & Point-in-time recovery, bitemporal correctness, materialized structural-unit versioning, event-bounded validity, and interpretive--statutory coherence & Event reification and bitemporal correctness \\
Causal opacity & Institutional provenance and duty of justification & Returns text or composed legal claims without the chain of acts that grounds them & Grounding-event attribution, lineage reconstruction, dual-chain surfacing, claim-level provenance traceability, and auditable chain composition & Event-level provenance, claim-level attribution, and deterministic interaction protocols \\
\bottomrule
\end{tabularx}
\end{table}

The same mapping can be stated as a minimal formal diagnostic. Let a legal retrieval substrate be represented as a knowledge graph $G=(V,E)$, where $V$ is a set of typed nodes, including stable legal references, textual fragments, structural-unit states, interpretive states, and legal events, and $E$ is a set of typed edges connecting them. Events are treated as first-class nodes. Where a relation is ternary, as in the transformation of a prior state into a subsequent state by an event, it is represented through reification.

The substrate exposes at least three families of relations and one derived retrieval function. The relations are $\operatorname{partOf}(x,y)$, for directed mereological containment, where $x$ is contained in, or structurally governed by, $y$; $\operatorname{stateOf}(s,r)$, linking a textual or interpretive state $s$ to the stable legal reference $r$ that it realizes; and $\operatorname{transforms}(e,s^{-},s^{+})$, for validity-affecting legal events that transform a prior state $s^{-}$ into a subsequent state $s^{+}$, with $s^{-}=\bot$ for originating events.

The derived function $\operatorname{statesAt}(r,T,c)$ returns the set of textual or interpretive states $s$ such that $\operatorname{stateOf}(s,r)$ and $s$ is valid at target legal time $T$ under legal context $c$. Its computation is deterministic and may rest on stored valid-time intervals attached to states, on traversal over the chain of $\operatorname{transforms}$ events, or on a combination of both. The function concerns valid time; bitemporal correctness further requires that valid time not be conflated with transaction time, that is, the time at which a state or event was recorded in the official information system. Here, $c$ denotes the relevant jurisdictional, institutional, or procedural context, and ``validity-affecting'' includes both textual changes and authoritative interpretive changes.

A provenance chain for a state $s_n$ is an ordered sequence
\[
\pi(s_n)=\langle(e_0,s_0),(e_1,s_1),\ldots,(e_n,s_n)\rangle
\]
such that $\operatorname{transforms}(e_0,\bot,s_0)$ and, for every $i>0$, $\operatorname{transforms}(e_i,s_{i-1},s_i)$. The sequence may include enactment, amendment, repeal, suspension, revival, correction, declaration of invalidity, or authoritative interpretation events, depending on the legal system represented. The target $s_n$ may be a textual state, an operative interpretive state, or a terminal/non-valid state. A state may have more than one relevant provenance chain where multiple authorities, parallel interpretive lines, or compound legal effects are represented.

Given a retrieval function $R(q,T,c)$ that returns an output $O$ for query $q$, target legal time $T$, and legal context $c$, the output is legally adequate within the scope of this framework only if it satisfies three closure conditions:

\begin{enumerate}[label=(\arabic*), leftmargin=2em, itemsep=0.5ex]
    \item \textbf{Structural-context closure:} for every returned textual fragment or structural-unit state in $O$, and for every returned interpretive state or material legal claim whose support depends on a provision, the governing structural context required for interpretation is recoverable. This includes ancestor relations and, where represented in the substrate, typed legal relations such as definitions, exceptions, and cross-references.
    
    \item \textbf{Temporal-contextual correctness:} for every returned textual version or operative interpretive state $s \in O$, there exists a stable legal reference $r_s$ such that $\operatorname{stateOf}(s,r_s)$ and $s \in \operatorname{statesAt}(r_s,T,c)$. If the query targets a legal reference $r$ for which $\operatorname{statesAt}(r,T,c)$ contains no state matching the query target, the system reports this absence rather than substituting the current, consolidated, or temporally adjacent state, and rather than inferring the state probabilistically from underlying instruments. Where multiple states are valid under distinct legal contexts, the system must distinguish them rather than collapse them into a single undifferentiated answer.
    
    \item \textbf{Provenance reconstructibility:} for every returned state $s \in O$, the system can recover an auditable provenance chain $\pi(s)$. For every material legal claim expressed in the synthesis layer over $O$, the system can recover the set of provenance chains $\{\pi_j\}_{j=1}^{k}$ supporting the states, interpretations, or authorities on which the claim depends.
\end{enumerate}

Mereological blindness is primarily the failure of condition~(1), diachronic blindness primarily the failure of condition~(2), and causal opacity primarily the failure of condition~(3). The formulation is deliberately lightweight: it is not a complete logic of law, but a diagnostic specification for retrieval outputs.

The three conditions form a checklist rather than a ranking. A system that satisfies two and fails the third may still be useful for narrow informational tasks, but it fails the stronger standard of validity-grounding legal retrieval developed here. The first two pathologies are failures of \term{state recovery}: synchronic in the case of mereological structure, diachronic in the case of temporal evolution. Causal opacity is different in kind: it is a failure of \term{exposure and auditability}. This distinction explains why the architectural responses developed in Section~\ref{sec:toward_neurosymbolic} differ: the blindness pathologies require substrate-level commitments that change what retrieval can recover, whereas causal opacity additionally requires output- and interface-level commitments that change what retrieval returns and what the user can audit.

Existing legal retrieval approaches are heterogeneous with respect to this framework. The claim is not that every system fails, but that the reviewed literature has not yet converged on a widely adopted and empirically validated legal-retrieval paradigm in which the three requirements are treated as co-constitutive. Section~\ref{sec:review} reviews the state of the art through this lens.

% =============================================================================
\section{Critical Review of Current Approaches}
\label{sec:review}

The literature on retrieval-augmented generation and legal knowledge representation does not ignore the commitments articulated in Section~\ref{sec:legal_domain}. For each of the three pathologies named in Section~\ref{sec:three_limitations}, identifiable lines of work address the corresponding requirement, sometimes with considerable sophistication. What the reviewed literature has not yet produced is co-satisfaction as a foundational design property. Existing strands typically foreground one condition, support another partially, and leave the remaining condition to downstream components, metadata conventions, or human verification. 

This section therefore reviews the literature by pathology rather than by technological category. Surveys of LLM--knowledge graph integration and temporal knowledge graphs map important parts of this terrain by treating each technology as an evolving family~\parencite{pan2024unifying,cai2024survey}. Our aim is narrower and more diagnostic: for each pathology, what progress exists, where it stops, and what prevents existing approaches from addressing the other two pathologies at the same time. Section~\ref{subsec:synthesis} closes with the synthesis argument that motivates Section~\ref{sec:toward_neurosymbolic}.

\paragraph{Relation to classical AI \& Law.}
The argument developed here should be distinguished from three established AI \& Law traditions. The first is rule representation, legal ontologies, and deontic reasoning, including LKIF-Core and LegalRuleML~\parencite{hoekstra2007lkif,palmirani2011legalruleml}, which asks how legal norms can be represented for reasoning. The second is case-based reasoning and precedent~\parencite{ashley2017analytics}, which models how courts reason from prior decisions. The third is legal information retrieval, exemplified by the COLIEE case- and statute-retrieval benchmarks~\parencite{goebel2024coliee}, which optimize relevance and entailment over legal texts. Our focus is prior to all three: how a retrieval substrate identifies which textual or interpretive state of a norm is available, valid, and institutionally grounded at a given time. For any of these computational approaches to operate safely over live, evolving legal corpora, they presuppose that the applicable rule has been retrieved at the correct version with auditable provenance, a presupposition that relevance-oriented legal IR does not by itself guarantee. Our framework specifies the retrieval-side conditions under which that presupposition holds. A distinct response to legal unreliability scales or specializes the language model itself, as in domain-adapted legal LLMs trained on large legal corpora~\parencite{colombo2024saullm,colombo2024saullm_large}. This is a complementary but orthogonal bet: greater in-domain model capacity improves fluency and legal-task performance, yet it does not supply the structural, temporal, and provenance guarantees that the three pathologies concern, which are properties of the retrieval substrate rather than of model scale.

The qualitative labels in Table~\ref{tab:literature_matrix} summarize the typical state of deployed or explicitly specified implementations in each strand as retrieval substrates, not the maximum theoretical capability of the underlying technology. The scale is ordinal and intended as a comparative diagnostic, not as a benchmarked empirical ranking.

\begin{table}[!ht]
\centering
\caption{Literature strands through the three-pathology lens.}
\label{tab:literature_matrix}
\footnotesize
\begin{tabularx}{\linewidth}{P{0.22\linewidth}P{0.13\linewidth}P{0.13\linewidth}P{0.13\linewidth}Y}
\toprule
\textbf{Strand} & \textbf{Mereology} & \textbf{Time} & \textbf{Provenance} & \textbf{Residual limitation} \\
\midrule
Legal document standards & Strong & Partial & Weak--partial & Structure, identifiers, and lifecycle metadata may be encoded, but traversal and event-chain retrieval are not enforced \\
Structure-aware chunking and multi-level retrieval & Partial--strong & Weak & Weak & Context can be enriched, but completeness still depends on routing and typed traversal beyond similarity \\
Graph and ontology RAG & Partial--strong & Weak--partial & Weak & Graph relations may be inferred or curated, but are not legal hierarchy, valid-time state, or provenance chains by default \\
Temporal and event KGs & Weak--partial & Strong & Partial & Temporal state is represented, but legal event semantics and retrieval protocols remain substrate-specific \\
Provenance modeling (PROV-O) & Weak & Partial & Strong & Vocabulary for provenance, not a retrieval-facing protocol that surfaces it \\
Explainable AI for legal applications & Weak & Weak & Weak (behavioural) & Explains model behaviour, not validity grounding \\
Tool-using agents$^\dagger$ & Substrate-dependent & Substrate-dependent & Partial (auditable trace) & Calls are auditable; orchestration remains probabilistic and tool semantics are inherited from the substrate \\
\bottomrule
\end{tabularx}
\vspace{0.5em}
\footnotesize\textit{Note}: $^\dagger$ Tool-using agents are an orchestration paradigm rather than a representation strand; their contribution to each pathology depends on the substrate exposed via tools.
\end{table}
\FloatBarrier

\subsection{Addressing Mereological Blindness}
\label{subsec:addressing_mereological}

Work on mereological preservation in legal retrieval has three broad strands: standards-based document structuring, structure-aware retrieval techniques, and ontology-grounded graph construction. Progress is real but uneven.

\paragraph{Document standards.}
Akoma Ntoso~\parencite{Palmirani2011,oasis2018akomantoso} and national document-markup initiatives such as Brazil's LexML XML Schema~\parencite{lexml2008xmlschema} represent mature formalizations of legislative structure. They encode part--whole hierarchy at fine granularity, including articles, paragraphs, items, and sub-items. Complementary identifier and metadata frameworks, such as the European Legislation Identifier~\parencite{eli_council_2012,eli_ontology} and LexML's URN specification~\parencite{lexml2008urn}, provide persistent identifiers and descriptive metadata for legal resources and, in some cases, fragments within them. Together, these standards address much of the representational variant of mereological blindness: at the data layer, the structural tree and stable references for its nodes are available.

What they do not solve by themselves is the operational variant. Standards specify how legal structure is encoded, identified, and exchanged; they do not prescribe how retrieval traverses it. In common RAG pipelines, the XML tree is serialized into retrievable units for lexical or vector indexing, while parent--child relations, sibling order, cross-references, and governing headings are omitted, stored as passive metadata, or left outside ranking and context assembly. The hierarchy is therefore not necessarily absent from storage; it is non-operative at retrieval time. Avoiding the operational variant therefore requires not only structurally valid source documents, but retrieval protocols that enforce typed traversal, ancestor closure, descendant completeness, and boundary-aware expansion.

\paragraph{Structure-aware retrieval and semantic chunking.}
A second line of work addresses the retrieval layer directly. Some semantic chunking strategies~\parencite{barnett2024seven} improve on fixed-size windows by respecting syntactic or discourse boundaries. In the legal domain, multi-layered embedding-based retrieval has been proposed for legislative texts~\parencite{lima2024multilayered}: legal materials are indexed at several levels of granularity, from whole documents down to articles, article components, and enumerative elements. Such approaches are relevant to mereological blindness because they recognize that enumerative provisions should not be embedded as isolated text; their representation should include, or be recoverable together with, the superior elements that govern them. Context-enrichment techniques follow the same intuition by prepending headings or superior provisions to dependent fragments. Generic multi-level indexing methods such as RAPTOR~\parencite{sarthi2024raptor} likewise construct hierarchical representations at different granularities; in legal corpora, analogous levels may correspond to items, articles, chapters, or higher structural units. A parallel line of work pursues domain-specialized dense retrieval, training or fine-tuning legal encoders to improve semantic matching over legal text~\parencite{li2023sailer}, with benchmark studies confirming that even strong dense retrievers and lexical baselines such as BM25 recover gold passages only partially on realistic legal queries~\parencite{zheng2025reasoning}. These models sharpen relevance estimation, but relevance is precisely the objective whose insufficiency for legal adequacy this paper argues: a better-calibrated similarity does not, by itself, enforce ancestor closure, point-in-time validity, or provenance.

These techniques mitigate important symptoms of the representational variant and go beyond naive local chunking. They do not, however, eliminate the pathology unless structural expansion is enforced as part of the retrieval protocol. Three limitations persist. First, enrichment is often local or bounded by a window and may miss definitions, exceptions, cross-references, or conditions located elsewhere in the statute. Second, multi-level indexing raises a routing problem: when a query is ambiguous about scope, which level should answer it? Third, vector similarity often remains the ranking function. Even with structurally coherent and context-enriched chunks, semantic proximity does not by itself enforce descendant completeness under scope queries or ancestor closure for dependent fragments~\parencite{reuter2025towards}. Recall of ``all sub-items of article $X$'' remains approximated rather than guaranteed unless deterministic structural traversal or typed graph expansion is added.

\paragraph{Graph-based and ontology-grounded approaches.}
A third line of work operates at the graph level. The \textit{Graph RAG} paradigm introduced by \textcite{edge2024graphrag} builds an LLM-derived graph index over a document collection, extracting entities, relationships, and claims from text chunks, applying community detection, and summarizing graph communities to support \textit{global sensemaking}. For open-domain corpora, this is a substantial advance over flat retrieval: the graph can expose relational structure that pure similarity search does not.

Applied to legal retrieval, however, the standard Graph RAG pipeline encounters a mismatch with the mereological structure that legal validity requires. Its communities are inferred from an entity--relation graph, whereas the structural units that matter legally --- titles, chapters, articles, paragraphs, items, and sub-items --- are formally decreed components of the legal instrument. An inferred community may cluster provisions that are topically related but mereologically unrelated, or split provisions that are mereologically bound. The graph reintroduces structure, but not necessarily the structure that legal validity requires.

Ontology-grounded variants address this mismatch by building the graph from a domain model rather than relying only on bottom-up graph induction. \textcite{sharma2025ograg} propose OG-RAG, which constructs a hypergraph representation grounded in a domain-specific ontology. For statutory retrieval, \textcite{louis2023finding} propose a graph-augmented dense retriever that incorporates the topological organization of legislation through a graph neural network. This is an important step toward structure-aware retrieval, although the graph improves ranking rather than enforcing deterministic ancestor closure or descendant completeness. For legal applications that must combine legal mereology with documentary identity and versioning, LRMoo~\parencite{lrmoo2024}, and before it FRBRoo, provide a useful upper-level modeling vocabulary because they distinguish abstract works from expressions and allow revised texts to be represented as new expressions. They should not, however, be read as legal-domain ontologies in themselves. Their role is to supply reusable bibliographic and cultural-heritage distinctions that legal retrieval architectures must specialize with legal concepts such as provisions, authorities, validity-affecting events, legal effects, and jurisdiction-specific relations.

The ontology-grounded strand is therefore one of the most principled responses to mereological blindness among the approaches reviewed here. Its limits are clear: it is costly to curate, and mereological correctness alone is not sufficient. A hierarchy-preserving graph, if not extended with temporal states and event provenance, remains silent about the next two pathologies.

\paragraph{Summary.}
Mereological blindness is the pathology for which the literature has traveled farthest, but most solutions remain stronger at representation than at retrieval-time enforcement.

\subsection{Addressing Diachronic Blindness}
\label{subsec:addressing_diachronic}

If mereological blindness is the pathology for which the literature has traveled farthest, diachronic blindness is the one least well integrated into mainstream RAG practice. Mature approaches to temporal representation exist, but they cluster in technical communities whose assumptions rarely enter legal RAG pipelines, and they address only portions of the full temporal structure articulated in Section~\ref{sec:legal_domain}. We organize the review around four paradigms commonly discussed in the temporal knowledge graph literature~\parencite{cai2024survey}: snapshot-based versioning, timestamped triples, event-centric modeling, and bitemporal modeling.

\paragraph{Snapshot-based versioning.}
The simplest model maintains separate graphs or documents for each state of the corpus: one snapshot per amendment, preserved in its entirety. This is the paradigm implicit in many institutional legislative repositories, where each new amendment triggers a newly published consolidated version. Snapshot-based versioning has one virtue: a past state is retrievable by selecting the appropriate snapshot. Its limitations are equally clear. First, storage and maintenance costs grow with the number of amendments and consolidations, which becomes burdensome for heavily amended codes. The problem is not only storage. When entire consolidated snapshots are indexed for retrieval, provisions that were not amended are repeated across successive versions, producing multiple identical or near-identical chunks for the same unchanged legal text. Unless the index deduplicates them or models structural-unit states explicitly, top-$K$ retrieval can be polluted by redundant copies rather than by legally distinct temporal states. Second, and more important for this paper, the relation between snapshots is not intrinsically represented. Even when Article~6 in the 1988 snapshot and Article~6 in the 2000 snapshot share a persistent identifier, the system need not represent the legal event that transformed one state into the other. Reconstructing the lineage therefore requires external alignment based on amendment metadata that snapshot-based versioning does not itself enforce. Point-in-time retrieval is possible; event-grounded temporal reconstruction is not guaranteed.

\paragraph{Timestamped triples.}
The second paradigm, dominant in much temporal knowledge graph work, extends the RDF triple from $(s,p,o)$ to $(s,p,o,[t_{start},t_{end}])$, attaching validity intervals to edges. In property graph implementations, the equivalent is a temporal property on a relationship. This model efficiently represents when a relation holds, and temporal query languages can retrieve the state of the graph at a past date~\parencite{snodgrass1999developing}. For legal knowledge, timestamped triples address valid time for relations but do not, by themselves, capture the institutional act that creates, modifies, or terminates those relations. The edge can state that it was valid during an interval; it does not necessarily state that Amendment~X opened that interval on date~$T$, that Ruling~Y superseded it on date~$T'$, or that a corrigendum altered the official record without changing the valid-time state. Such information can be added through named graphs, reification, or provenance annotations, but it is then no longer supplied by the timestamped-triple paradigm alone. Unless retrieval is designed to surface those annotations, provenance remains external to the returned temporal state. For high-stakes legal applications in which each returned provision must be traceable to an act, this is insufficient.

\paragraph{Event-centric models.}
The third paradigm, represented in the broader KG literature by EventKG~\parencite{gottschalk2018eventkg} and formalized earlier by the Simple Event Model~\parencite{vanhage2011design}, reifies events as first-class nodes rather than treating them only as edge properties. A legislative amendment, in this paradigm, becomes an Event node with its own properties, such as date, authority, type, and source, and with typed relations to the provisions it creates, modifies, suspends, revokes, or interprets. 

Applied to law, this model was anticipated by~\textcite{lima2008ontology} in their FRBRoo-based treatment of legal resources. The legal contribution of that line of work is to separate the stable identity of a legal resource from its time-indexed textual expressions, and to model legal change as the production of new resource states through identifiable events. LRMoo~\parencite{lrmoo2024}, as the successor vocabulary to FRBRoo in the IFLA/CIDOC family, updates this modeling path. It is not a complete legal-domain ontology, but an upper-level vocabulary for identity, expression, and versioning. Recent legal-domain work specializes this LRMoo-based pattern for component-level legal versioning and event-centered reconstruction of legal norms~\parencite{demartim2025diachronic}.

The event-centric paradigm is the most natural fit for legal dynamism among the paradigms reviewed here. It honors the operational-closure argument developed in Section~\ref{sec:legal_domain} by making the legally recognized trigger of validity transitions a first-class, queryable entity. It can handle provenance natively when the substrate exposes event relations to retrieval: to ask what act grounds the current text of a provision is to traverse from the returned state to the event or sequence of events that produced it. It also accommodates the multiplicity of termination and transformation causes, including express repeal, tacit repeal recognized by authority, expiration, suspension, correction, declaration of invalidity, and authoritative interpretation, by typing events rather than forcing all temporal change into a single schema.

Legal-domain instances remain comparatively sparse in the reviewed literature. Work exists at the level of proposals and prototypes, while most operational systems discussed in the literature continue to rely on document-level snapshots, consolidated texts, or external version-control metadata. The claim is not that no institutional or proprietary system represents legal events; rather, no widely documented and empirically evaluated legal-retrieval paradigm yet treats events as the primary unit of temporal modeling in a production-grade legal KG. Recent proposals begin to address this combination explicitly, integrating event reification with bitemporal modeling for legal corpora; we discuss representative instances in Section~\ref{subsec:instances}.

\paragraph{Bitemporal modeling.}
The fourth paradigm formalizes orthogonality between \textit{valid time} and \textit{transaction time} (Section~\ref{sec:legal_domain}), allowing each dimension to be queried independently~\parencite{snodgrass1999developing}. For law, this is not a luxury. It is required to handle \textit{vacatio legis}, where a provision is published on one date but becomes legally effective on a later one; retroactive provisions, where a law enacted on one date applies to events that occurred earlier; delayed or conditional effectiveness; and corrigenda or republications, where the official record changes without necessarily changing the valid-time state. A unitemporal model collapses these distinctions and produces systematic errors precisely at the edges where temporal accuracy matters most. Bitemporal modeling is standard in temporal databases and appears in parts of the TKG literature. In legal retrieval systems, however, it is rarely exposed as a first-class retrieval capability. A common practice is to record a single date stamp per provision and treat it as both the date of official registration and the date of legal effect. That simplification is correct for many ordinary cases and wrong for precisely the cases that most require formal care: \textit{vacatio legis}, retroactivity, corrigenda, and delayed effectiveness.

\paragraph{The gap of interpretive time.}
The four paradigms above are not intrinsically limited to legislative time. In principle, event-centric and bitemporal models could represent judicial or administrative interpretation as validity-affecting events. As typically imported into legal retrieval architectures, however, they are applied primarily to textual legal time: how statutes and regulations evolve through authored amendments, consolidations, and recorded validity intervals. They rarely treat \term{interpretive time} as a first-class retrieval dimension: the dimension along which binding judicial or administrative authorities alter the operative norm without altering the legal text. The gap is therefore architectural rather than conceptual. Legal retrieval systems usually model the statutory or regulatory text as the primary temporal artifact, while case law and administrative interpretation are treated, if at all, as external overlays. For civil-law jurisdictions where constitutional adjudication or binding precedent can alter operative meaning without textual amendment, and for common-law jurisdictions where the relation between statute and case law is even more entangled, this gap is not a detail. As the Brazilian presumption-of-innocence example in Section~\ref{subsec:diachronic_blindness} illustrates, an operative norm can reverse without a single textual change. Interpretive time is therefore a substantial fraction of diachronic blindness, and one that temporal-representation techniques can in principle model, but that legal retrieval architectures have only begun to expose as a retrieval-facing capability.

\paragraph{Summary.}
Each paradigm addresses a facet of diachronic blindness. Snapshots handle point-in-time retrieval coarsely but do not guarantee event-grounded lineage. Timestamped triples represent temporal intervals efficiently but do not, by themselves, expose the legal acts that open, modify, suspend, interpret, or close those intervals. Event-centric models represent validity-affecting institutional events, but remain uncommon as legal-retrieval substrates. Bitemporal modeling captures valid-time versus transaction-time orthogonality, but rarely combines with event reification and retrieval-facing interfaces in practice. Legally adequate temporal handling requires their combination: point-in-time recovery, structural-unit versioning, event-bounded validity, bitemporal orthogonality, and interpretive--statutory coherence. Among the paradigms reviewed here, event-centric bitemporal modeling is the most promising architectural direction because it can connect temporal states to the institutional acts that produce them while keeping valid time and transaction time distinct. It is not sufficient on its own, however: it must be coupled to a typed legal substrate and retrieval protocols that expose those states and events as auditable outputs. Joint instances in widely deployed, documented, and empirically evaluated legal retrieval systems remain limited.

The broader reason is partly disciplinary: temporal KGs and modern RAG pipelines have advanced largely in parallel. Legal RAG systems inherit retrieval architectures not designed for legal time, while temporal-KG research has not generally centered on retrieval-facing interfaces that downstream legal AI systems can query natively. Closing this gap remains an architectural challenge.

\subsection{Addressing Causal Opacity}
\label{subsec:addressing_causal}

Causal opacity is addressed only indirectly by the literatures most often invoked in response to it: explainable AI, provenance modeling, and tool-using agents. None of these strands, by itself, targets causal opacity in the legal sense defined in Section~\ref{subsec:causal_opacity}: the requirement that a retrieval output expose the institutional chain of acts that grounds the returned legal state or synthesized claim. Each strand nevertheless contributes components that a legally adequate approach would have to combine.

\paragraph{Explainable AI and its limits.}
The XAI literature has produced a sophisticated toolkit for behavioural explanation: attention visualizations, feature attribution, counterfactual probing, influence scoring, and related methods. Applied to the legal domain~\parencite{richmond2024explainable}, these techniques address questions such as: \textit{why did the model produce this output, given its inputs, parameters, retrieved context, or local decision boundary?} This is a real contribution, but it is different from the question that causal traceability, in the legal sense used here, requires: \textit{which institutional acts ground the legal state or claim asserted by this output?} Model introspection cannot by itself answer the second question, because the answer is not located in the model's internal behaviour. It is located in the institutional record of enactments, amendments, repeals, rulings, administrative acts, and other legally recognized events that the retrieval substrate must surface.

The distinction matters in practice. An XAI-augmented RAG system may produce rich visualizations of which tokens attended to which tokens, which retrieved fragments most influenced the generated answer, or which features drove a similarity or classification score. A legal practitioner reviewing such an output may learn something about the behaviour of the system, but still be unable to audit the legal grounding of the answer. The visualization does not identify \textit{which amendment introduced the clause}, \textit{which ruling fixed its current interpretation}, or \textit{what validity chain supports the final claim}. Behavioural transparency and institutional traceability are therefore not the same thing. The former concerns the system's path to an output; the latter concerns the legal order's path to the norm or interpretation that the output asserts. For legal adequacy, the second is indispensable.

\paragraph{Provenance modeling.}
The provenance literature, formalized in W3C PROV-O~\parencite{lebo2013prov}, provides a vocabulary for expressing which entities, activities, and agents participated in the production of a resource: precisely the semantic shape needed for a legal event chain. A PROV-O trace for a provision could model each amendment or other validity-affecting act as a \texttt{prov:Activity}; the prior and subsequent provision states as \texttt{prov:Entity} instances; the legislative or judicial authority as a \texttt{prov:Agent}; and the source act, prior version, and generated version through relations such as \texttt{prov:used}, \texttt{prov:wasGeneratedBy}, \texttt{prov:wasDerivedFrom}, \texttt{prov:wasAssociatedWith}, and temporal properties such as \texttt{prov:startedAtTime}, \texttt{prov:endedAtTime}, or \texttt{prov:generatedAtTime}. This is close to the representational pattern that event-centric temporal KGs instantiate, whether or not they explicitly cite PROV-O.

Provenance modeling provides a representational vocabulary, not a retrieval-facing protocol. A system can store PROV-O-compliant provenance for every provision and still return only the current text at query time, leaving the event chain latent in the data layer. Closing this gap does not require a natural-language agent, but it does require some interface --- an API, query protocol, UI affordance, or deterministic primitive --- that exposes provenance as a first-class retrievable object rather than optional metadata. This is not a default property of general-purpose RAG systems.

\paragraph{Tool-using agents.}
A promising architectural direction for causal transparency comes from the agentic literature. The ReAct paradigm~\parencite{yao2023react} and related tool-using agent proposals~\parencite{schick2023toolformer} decompose retrieval from a single black-box step into a sequence of explicit actions: plan, invoke a tool, observe the result, plan again, invoke again. Each tool call is logged. The trace of calls is, in principle, an operational analogue of the institutional causal chain: a record of which retrieval action produced which content, even if not yet of which legal act produced which legal state.

For legal retrieval, the agentic paradigm is attractive because it aligns tool calls with the kinds of operations a legal analyst actually performs: resolve an identifier, perform point-in-time retrieval, traverse a provision's lineage through its amendments and interpretations, and compare versions textually. If these operations are exposed as deterministic tools and each agent action is logged, the operational trace is auditable by construction.

Four limitations qualify this promise. First, when tool-using agents are connected to structured stores, they often rely on the agent generating queries in a general query language, such as SQL, Cypher, or SPARQL. This introduces a new failure mode: generated queries can be syntactically valid but legally incorrect, for example, by omitting the validity filter that excludes repealed versions. Empirical studies of text-to-SQL~\parencite{pourreza2023dinsql} show meaningful error rates even for carefully engineered systems; for legal retrieval, where silent errors are unacceptable, unconstrained dynamic query generation is a poor foundation. Second, the tool set exposed to an agent is often generic rather than composed of the domain-specific primitives that legal retrieval requires: point-in-time recovery, lineage traversal, impact analysis, and provenance-chain reconstruction. Third, the trace of tool calls is auditable but not yet tied to institutional legal events: logging that the agent called \texttt{getVersion(id, date)} is not the same as logging which amendment, ruling, or administrative act instituted the returned state. Full causal traceability requires the tools themselves to return event-annotated outputs, which in turn requires the underlying KG to be event-centric.

A fourth limitation persists even when the preceding three are addressed. Suppose an agent is equipped with a domain-specific, event-annotated tool set; if orchestration is left to a generic probabilistic planner, the system must still decide which tool to call, in what order, and with what inputs. Multi-hop legal queries expose this residual vulnerability. Answering ``what norm applied to legal situation type $S$ on date $T$?'' may require composing several operations: retrieve the version valid at $T$ under the relevant legal context, check whether any binding judicial or administrative authority had altered its operative interpretation by $T$, and check whether any authoritative recognition event had superseded the relevant textual or interpretive state, while flagging possible tacit supersession where the transition depends on legal interpretation. Omitting any one of these steps can silently produce the wrong answer.

Tool correctness alone does not prevent orchestration error: a probabilistic planner can invoke valid tools in an incomplete or incorrect sequence, and the output will be plausible, fluent, and legally wrong. This observation strengthens, rather than undermines, the case for tool-using agents in legal retrieval, but only when tool use is constrained by domain-specific protocols. The auditability gains must be paid for by constraining the agent's orchestration space, either through fixed protocols that encapsulate the temporal and provenance logic, verification layers that check required steps, or constrained planning models paired with legal workflow validators. Generic ReAct-style loops over generic or underspecified legal tools are not sufficient.

\paragraph{Summary.}
XAI, provenance modeling, and agentic architectures each supply a component of a legally adequate response to causal opacity, but none is complete alone. XAI explains system behaviour, not institutional grounding; PROV-O supplies a provenance vocabulary, not a retrieval function that surfaces it; tool-using agents provide auditable execution traces, but those traces become legally meaningful only over event-centric substrates whose tools return event-annotated outputs. The integration required to answer provenance questions natively is achievable, but remains a research and engineering challenge rather than settled practice.

\subsection{Synthesis: Why Partial Solutions Do Not Compose}
\label{subsec:synthesis}

The review above might suggest a straightforward integration recipe: combine the strongest approach to each pathology and solve them jointly. Use ontology-grounded graph construction for structure, event-centric bitemporal modeling for time, and provenance-aware interfaces for grounding; wire them together. The recipe is attractive, but insufficient when applied as a retrofit over substrates designed for other purposes. We identify three reasons.

First, the approaches are built around different primary units and invariants. Ontology-grounded graph construction assumes a graph organized by domain concepts or structural units, often treated as stable within a retrieval session~\parencite{sharma2025ograg}. Event-centric temporal modeling assumes that legal states have lifecycles driven by first-class events. Tool-using or API-based retrieval assumes a set of callable operations with contracts stable enough for planning, logging, and audit. These assumptions are not incompatible in principle; indeed, the next section argues that they must be integrated. The problem is that they do not compose automatically when added after the fact. Retrofitting the three strands onto a single retrieval stack tends to produce impedance mismatches at the boundaries: structural nodes are not necessarily temporal states; temporal events are not necessarily exposed as retrieval outputs; and tools are only as legally meaningful as the substrate and contracts they encapsulate.

Second, partial solutions can mask or amplify the pathologies they do not address. Adding hierarchical structure to a graph without temporal handling may produce structurally rich but temporally incorrect outputs, which can be legally riskier than obviously incomplete outputs because their apparent precision invites trust while their temporal correctness remains difficult to verify. Adding temporal handling without retrieval-time structural context produces a different failure: the system may correctly reconstruct the past version of a fragment whose legal meaning depends on ancestors or on represented legal relations, such as definitions, exceptions, or cross-references, that are not reconstructed with the same care. Adding an auditable agentic trace without event-level provenance creates yet another failure: the system logs which tools were called, but not which institutional acts grounded the returned legal state. In these configurations, partial solutions do not merely leave residual pathologies untouched; they can make those pathologies harder to detect.

Third, and most fundamentally, the three pathologies are not independent specifications that can be satisfied in parallel. They are, as argued at the end of Section~\ref{sec:three_limitations}, jointly constitutive of validity-grounding legal retrieval. A retrieval output is legally adequate for this purpose only when the returned content is structurally contextualized, temporally correct, and causally traceable. An output that fails one of these conditions may still be useful for narrow informational tasks, but it does not qualify as legally adequate for high-stakes legal retrieval as defined in this paper. Nor does a system that routinely produces such outputs qualify as an adequate retrieval substrate for that purpose. The right architectural move is therefore not to add capabilities one at a time, but to design a substrate in which structure, time, and provenance are co-specified from the start.

Section~\ref{sec:toward_neurosymbolic} argues that substrates designed for such co-satisfaction, when coupled to language-model interfaces or agentic orchestration layers, fit naturally within the neuro-symbolic paradigm. It identifies a specific set of architectural commitments that make joint satisfaction possible. The three pathologies are therefore not a loose list of research problems. They are a single diagnostic checklist whose items cannot be fully addressed in isolation without sacrificing the architectural coherence required for legal adequacy.

% =============================================================================
\section{Toward Neuro-Symbolic Legal AI}
\label{sec:toward_neurosymbolic}

The review in Section~\ref{sec:review} presented the three pathologies as diagnoses of a fragmented literature. A reasonable reading would be that each pathology is a technical gap awaiting a technical fix, and that integration is a matter of engineering discipline. We argue that this reading is incomplete. The deeper issue is not merely technical but paradigmatic. Retrieval in the dominant RAG paradigm is conceived as a problem of \textit{approximating relevance}: finding content statistically close to what the user asked for. Legal retrieval, as defined by the three commitments of Section~\ref{sec:legal_domain}, is a problem of \textit{reconstructing validity}: establishing which content is formally grounded, at what time, and by what institutional act. These tasks have different success criteria, failure modes, and epistemic warrants. A paradigm calibrated for approximate relevance cannot reliably satisfy validity reconstruction by tuning similarity alone or by adding isolated post-hoc components.

The architectural move required is to design for determinism where the legal domain demands determinism, and to reserve probabilistic inference for the layers where it legitimately belongs. This is the commitment made by what we call \term{deterministic-by-design} legal retrieval. The term does not mean that every legal question has a mechanically determinate answer, nor that deterministic systems eliminate interpretive disagreement. It means that, within the represented scope, legality-critical retrieval operations such as structural traversal, point-in-time recovery, and provenance-chain reconstruction should be reproducible, auditable, and governed by explicit legal relations rather than inferred from semantic proximity.

This architectural posture fits naturally within the broader vocabulary of neuro-symbolic AI, especially the literature on coupling large language models with knowledge graphs and other symbolic substrates~\parencite{pan2024unifying}. The terms, however, are not identical. \textit{Neuro-symbolic} names the general combination of neural and symbolic components. \textit{Deterministic-by-design} names the legal-domain requirement that validity grounding, temporal correctness, and institutional provenance be handled by controlled symbolic or procedural mechanisms. A deterministic legal retrieval substrate can satisfy the framework without a language model or agent; it becomes neuro-symbolic when a neural component is added for natural-language understanding, planning, or synthesis on top of that substrate.

The rest of this section makes the architectural commitment concrete. Subsection~\ref{subsec:commitments} identifies four commitments that follow from jointly addressing the three pathologies. Subsection~\ref{subsec:instances} reviews existing work that instantiates, approximates, or supports these commitments. Subsection~\ref{subsec:tradeoffs} addresses the costs and limits of this direction.

\subsection{Architectural Commitments}
\label{subsec:commitments}

The preceding synthesis motivates a positive architectural specification. If the three pathologies are jointly constitutive of validity-grounding legal retrieval, then addressing them requires more than adding isolated capabilities to an existing RAG pipeline. It requires a small set of design commitments fixed at the substrate level. We name four. Each is individually insufficient; together, they define the space of deterministic-by-design legal retrieval architectures.

\paragraph{C1. Ontological primacy.}
Structure must be a first-class property of the retrieval substrate, not merely a metadata annotation layered on top of text. An architecture satisfies ontological primacy when its primary data objects are legal-domain entities, such as normative acts, stable legal references, structural legal units, and versioned states of those units, rather than text fragments with structural labels. The distinction is consequential. In an annotation-based design, text is primary and structure is descriptive; the retrieval function reaches structure only by inspecting metadata attached to text. In an ontology-primary design, structure is the substrate on which text hangs: the retrieval function traverses typed legal relations directly, and text is attached to specific ontological nodes. Ontological primacy is therefore the architectural response to the symmetry mismatch identified in Section~\ref{subsec:mereological_blindness}: it replaces similarity among fragments with \term{typed structural traversal} over legally asymmetric relations.

This commitment addresses mereological blindness at its representational root. A retrieval function that operates over ontological entities can enforce scope integrity, support completeness under hierarchical queries, and recover governing context by traversing the part--whole relation directly. A function that operates over text fragments with structural labels can approximate these behaviours, but the labels remain inputs to the function rather than constraints on its outputs. Ontology-grounded RAG approaches~\parencite{sharma2025ograg} move toward this commitment by construction; standard Graph RAG~\parencite{edge2024graphrag}, when built through bottom-up extraction alone, does not satisfy it by default.

\paragraph{C2. Event reification.}
Validity transitions must be represented as first-class, queryable entities, not merely as passive edge annotations or timestamps. An architecture satisfies event reification when every represented legislative act, judicial ruling, or other validity-affecting event exists in the graph as an entity with its own properties --- date, authority, type, affected provisions --- and its own relations to the states it creates, modifies, suspends, interprets, or terminates~\parencite{vanhage2011design,gottschalk2018eventkg}.

This commitment addresses diachronic blindness structurally rather than by accumulation. A graph in which events are only timestamps on edges can answer \textit{when} a relation held; a graph in which events are queryable entities can answer \textit{which act} made the relation hold, \textit{which subsequent act} terminated or superseded it, and \textit{how} the chain of acts composes. Event reification also provides the natural hook for extending the paradigm to interpretive time. A binding ruling can be modeled architecturally as the same kind of entity as a legislative amendment: a typed event that creates or fixes an operative legal state. The graph therefore accommodates interpretive change without redesign, provided that interpretive events are included in the represented scope.

\paragraph{C3. Bitemporal correctness.}
Valid time and transaction time must be independently representable and independently queryable. An architecture satisfies bitemporal correctness when it represents each versioned textual or interpretive state with two distinct temporal intervals: the interval during which the state produces legal effect (valid time) and the interval during which the state is recorded in, or known to, the official information system (transaction time)~\parencite{snodgrass1999developing}.

This commitment addresses the edge cases that unitemporal modeling systematically mishandles: \textit{vacatio legis} periods during which a published norm has not yet taken effect; retroactive provisions that apply to acts performed before their enactment; delayed or conditional effectiveness; and corrections or erratas that adjust the recorded state without necessarily adjusting the legal state. These cases are not exotic. They are the load-bearing cases where correctness matters most. A system that collapses the two times handles the majority of ordinary queries correctly and the consequential minority incorrectly: a failure mode worse than uniform inaccuracy because it is invisible on standard benchmarks.

\paragraph{C4. Deterministic interaction protocol.}
The retrieval-facing interface to the knowledge substrate must expose domain-specific, semantically typed primitives, not rely on open-ended query generation as the default access mode. An architecture satisfies this commitment when access to the substrate is mediated by a fixed, auditable set of functions --- point-in-time retrieval, lineage traversal, impact analysis, version comparison, provenance-chain reconstruction --- each with a precise contract about inputs, outputs, temporal semantics, and provenance.

The contrast is with two alternative interfaces. The first is dynamic query generation: a model or agent produces queries in SQL, Cypher, or SPARQL, which are then executed against the graph. This interface inherits the reliability limits of text-to-code models~\parencite{pourreza2023dinsql} and, for legal retrieval, introduces silent failure modes: queries that execute successfully but omit validity filters, mishandle temporal constraints, or traverse the wrong legal relation. The second is raw vector retrieval: the system issues an embedded query, receives top-$K$ fragments, and reasons over them. This interface inherits the completeness and grounding limits critiqued throughout Sections~\ref{sec:three_limitations} and~\ref{sec:review}.

A domain-primitive interface addresses both. By fixing the operations available, it reduces the space of malformed queries and constrains legal error to typed, auditable failure modes. By typing inputs and outputs ontologically, it makes each retrieval action an auditable step in a reconstructible chain. By encapsulating temporal and provenance logic inside primitives, it relieves downstream agents or applications of the burden of re-deriving that logic at every query. The calling system plans or requests; the primitive executes; the trace becomes available for audit. This is the architectural analogue of the causal-traceability commitment of Section~\ref{subsec:causality}.

When such a substrate is coupled to a language model, this commitment clarifies the model's role. The model is not eliminated; it is repositioned. It translates the user's natural-language intent into a plan of primitive calls and synthesizes the primitives' outputs back into prose. It does not decide what the law says on a date, compute validity, or reconstruct provenance. Neural components handle linguistic interpretation, planning, and synthesis; symbolic or procedural components handle validity grounding, temporal resolution, and institutional traceability. The architecture is \textit{neuro-symbolic} only at this coupled level: the deterministic substrate can exist without the neural interface, but the neural interface becomes legally reliable only by operating over such a substrate.

\paragraph{Joint satisfaction.}
The four commitments are mutually reinforcing rather than merely additive. C1 gives C2 stable legal states to attach to; C2 gives C3 event-level provenance by identifying the acts that open, close, or transform temporal intervals; C3 gives C4 determinate temporal semantics for primitives such as ``retrieve the version valid at date~$T$''; and C4 turns the first three commitments into retrieval capabilities rather than merely stored structure. This dependency explains why partial adoption tends to disappoint: C4 without C2 exposes primitives without event-level outputs, while C1 and C3 without C2 provide structure and time without a provenance thread connecting them. The commitments form a single architectural package.

\subsection{Existing Instances}
\label{subsec:instances}

The four commitments are not merely aspirational. The preceding review has already identified strands of work that instantiate, approximate, or provide the technical apparatus for each commitment separately; the point here is narrower. The commitments are technically available, even if their legal-domain co-satisfaction remains rare.

Ontological primacy (C1) is approached by ontology-grounded RAG systems~\parencite{sharma2025ograg} and by structure-aware statutory retrieval models~\parencite{louis2023finding}. Legal-domain models that use LRMoo~\parencite{lrmoo2024} or its FRBRoo predecessor~\parencite{frbroo2016} as an ontological backbone --- as in the early adaptation of FRBRoo to legal resources by \textcite{lima2008ontology} --- move toward this commitment by making works, expressions, and related legal-document entities first-class. These models do not by themselves provide a complete legal-domain ontology, but they supply upper-level distinctions that legal retrieval architectures can specialize for provisions, structural units, versioned states, authorities, and validity-affecting events.

Event reification (C2) has mature analogues in the general knowledge graph literature, notably EventKG~\parencite{gottschalk2018eventkg}, and a formal precedent in the Simple Event Model~\parencite{vanhage2011design}. Documented legal-domain work is comparatively sparse, but the modeling pattern is clear: amendments, repeals, rulings, corrigenda, and other validity-affecting acts must be represented as queryable entities rather than as passive timestamps or textual annotations.

Bitemporal correctness (C3) is standard in temporal databases~\parencite{snodgrass1999developing}, but its integration with event-centered legal retrieval is still uncommon in AI systems. Recent property-graph work on legislative knowledge management provides a partial legal-domain approximation: it models laws and articles as graph nodes, represents amendment and abrogation relations among them, and supports valid-time retrieval of the version of a law in force at a given timestamp~\parencite{colombo2025legislativepg}. Such work approximates C1, C2, and the valid-time component of C3, although it does not by itself provide the deterministic, semantically typed retrieval protocol that C4 requires.

Deterministic interaction protocols (C4) emerge at the intersection of tool-using agents, domain-specific APIs, and institutional legal information systems. The general agentic literature~\parencite{yao2023react,schick2023toolformer} establishes the pattern of decomposing tasks into explicit tool calls. The legal-domain requirement is stricter: the operations exposed to downstream systems must be semantically typed primitives that encapsulate structural, temporal, and provenance logic behind stable contracts. Individually, each commitment is already evidenced by the third-party work reviewed above, which jointly reaches C1--C3 in the property-graph legislative modeling noted earlier. Recent work demonstrates the integration of all four within legislative and constitutional retrieval~\parencite{demartim2025graphrag,demartim2025primitiveapi}, serving as an existence proof that the four commitments are jointly implementable---not as evidence that the framework is satisfied in full, which no widely documented system yet establishes.

The takeaway is that none of the four commitments is speculative. Each has precedents, approximations, or partial instantiations. What remains rare is their co-satisfaction in widely adopted, documented, and empirically evaluated legal-retrieval architectures.

\subsection{Trade-offs and Honest Limitations}
\label{subsec:tradeoffs}

The deterministic-by-design direction is not free, and we do not claim otherwise. Four limitations deserve explicit acknowledgment. A general caveat frames all four: determinism guarantees reproducibility and constraint enforcement conditional on a correct substrate; it does not guarantee substantive legal truth, nor does it eliminate errors in ontology design, source ingestion, doctrinal classification, or jurisdiction-specific interpretation.

\paragraph{Curation cost.}
Ontology-primary substrates require substantial upfront investment. They are not built by running an extraction pipeline over PDFs; they require legal expertise to map documents onto an ontology, validate the mapping, and maintain alignment as amendments, rulings, corrections, and interpretive events arrive. This cost is real and scales with jurisdictional coverage. Against it must be set the cost of the alternative: systems that produce fluent but anachronistic, structurally incomplete, or institutionally ungrounded outputs in a domain where those failures have legal consequences. The comparison is therefore not between an expensive correct system and a cheap correct system, but between a costly, auditable system whose correctness depends on curated structure and a cheaper, under-constrained system whose errors are externalized onto downstream users.

\paragraph{Scope limits.}
The paradigm applies to \textit{quaestio juris}: which norms are available, in what version, and with what provenance. It does not extend to other dimensions of legal practice, including \textit{quaestio facti} --- whether real-world events occurred, whether witnesses are credible, or whether evidence establishes a state of affairs --- nor to judicial discretion, evaluative judgment on open-textured concepts, or other operations that act on legal material once it has been identified. A legal retrieval architecture that satisfies the four commitments provides the normative substrate over which these subsequent operations are performed; it does not perform them. Extending deterministic guarantees from norm identification to factual reconstruction, evaluative judgment, or discretionary application would be both practically unjustified and ethically hazardous.

\paragraph{Jurisdictional generalization.}
The architectural commitments are jurisdiction-agnostic; their instantiations are not. Drafting conventions, event taxonomies, authority hierarchies, and validity semantics vary across jurisdictions. A deterministic-by-design system therefore does not generalize by pure composition; it generalizes by principled adaptation. Upper-level ontologies such as LRMoo~\parencite{lrmoo2024} provide reusable invariants, including the distinction between works and versioned expressions and the separation between identity and state. What requires jurisdiction-specific authoring is the lower layer: local hierarchy, event subtypes, competent authorities, and the legal effects of different acts. The adaptation cost is meaningful but bounded.

\paragraph{Residual tensions.}
A final limitation concerns the apparent trade-off between flexibility and correctness. A system constrained to deterministic primitives cannot answer outside its substrate as readily as a probabilistic retriever can. It may decline, return empty, or flag an unsupported query. This trade-off is real, but under legal conditions its direction is asymmetric. Probabilistic systems fail fluently: they produce plausible outputs that may be anachronistic, structurally incomplete, or institutionally ungrounded. Deterministic systems are designed to fail explicitly: they return a warranted state with an audit trail, report that no represented state satisfies the query, or flag that the query falls outside the substrate's coverage or rules. The paradigm does not eliminate errors; it relocates them to points where they are more detectable and auditable.

The paradigm is therefore not proposed as a universal replacement for probabilistic retrieval in legal contexts. It is proposed as the architectural basis for applications in which legal adequacy, as defined by the three domain-level commitments, is non-negotiable: applications for which silent failure is intolerable and scoped refusal is acceptable. For exploratory research, informal advice, or general-interest legal information, probabilistic retrieval may remain appropriate. For high-stakes legal applications where outputs may ground consequential decisions, the asymmetric cost of error argues for the deterministic direction.

% =============================================================================
\section{Research Agenda}
\label{sec:agenda}

The framework and architectural commitments developed above establish the outlines of legally adequate retrieval, but they do not exhaust the research questions that this direction opens up. This section identifies four thematic fronts on which future work is likely to concentrate. We frame them thematically rather than by pathology or by commitment because each front cuts across the diagnostic categories: interpretive time, evaluation methodology, cross-jurisdictional portability, and human--system collaboration.

\subsection{Interpretive Time and the Doctrinal Dimension}
\label{subsec:agenda_interpretive}

Section~\ref{subsec:addressing_diachronic} identified interpretive time as the facet of diachronic blindness least integrated into retrieval architectures. The statutory layer of temporal modeling, however imperfect in practice, has relatively well-understood formal foundations: enactment, amendment, repeal, validity intervals, and transaction-time records. The interpretive layer has rich doctrinal and AI \& Law traditions, especially in work on precedent and case-based reasoning, but those traditions have not yet been integrated into retrieval architectures in the same operational form. Binding judicial or administrative decisions can create, fix, modify, suspend, or terminate operative states of provisions whose text remains unchanged. Treating those decisions as first-class validity-affecting events remains an open architectural problem.

Four questions organize this front. First, what formal ontology distinguishes binding from non-binding interpretive events? Jurisdictions vary widely: some treat appellate decisions as precedent, while others reserve binding force for specific procedural classes, such as constitutional court rulings, decisions with \textit{erga omnes} effect, administrative interpretations, or consolidated lines of authority. A general framework must accommodate these variations without collapsing them into a single undifferentiated \textit{case law} category.

Second, how should the temporal effect of interpretation be represented? Interpretive events may operate prospectively, retroactively, only between the parties, \textit{erga omnes}, or with expressly modulated effects. A retrieval system that asks what operative norm governed a date $T$ must therefore know not only that an interpretive event occurred, but when and for whom it altered the operative legal state.

Third, how should tacit overruling be represented? A later ruling may not explicitly reverse a prior decision, but may adopt reasoning incompatible with it. Tacit overruling is as consequential as express reversal, but harder to capture formally because the inference that a new ruling supersedes an old one is itself a matter of doctrinal interpretation. A deterministic retrieval system should not infer such transitions silently; it must either rely on an authoritative recognition event or flag the transition as requiring legal interpretation.

Fourth, how should jurisdictional or intra-court divergence be represented when two panels, chambers, or courts simultaneously maintain inconsistent interpretations of the same provision? A bitemporal model with a single valid state per provision cannot represent such divergence without distortion. The model must accommodate parallel operative states indexed by legal context: court, chamber, procedure, territorial scope, precedential force, and affected parties.

Progress on this front will likely come from connecting two traditions: event-centric temporal modeling and AI \& Law research on case-based reasoning, precedent, and doctrinal change. The productive move is to represent interpretive authorities as typed legal events, model their temporal and jurisdictional effects explicitly, and expose those effects through retrieval protocols capable of answering point-in-time questions about operative legal meaning.

\subsection{Evaluation Methodology for Deterministic Retrieval}
\label{subsec:agenda_evaluation}

Current benchmarks for legal retrieval, including LegalBench-RAG~\parencite{pipitone2024legalbench} and related legal-RAG benchmarks, evaluate retrieval largely within a relevance paradigm inherited from general information retrieval. Even when they move beyond document IDs to precise legal text segments, as LegalBench-RAG explicitly does, the core question remains: did the system retrieve the gold-standard relevant material, and how highly was that material ranked? Precision, recall, ranking metrics, and segment-level overlap are valuable for measuring textual relevance. They do not, by themselves, measure whether the retrieved material is structurally complete, temporally valid, or institutionally grounded.

More recent evaluation frameworks for RAG pipelines, including RAGAS-style metrics~\parencite{es2024ragas} and related LLM-as-judge proposals, do not resolve this mismatch. They refine the measurement of whether a generated answer is faithful to the retrieved context, whether the context is relevant to the question, and whether the answer appears correct relative to that context. These are useful metrics, but they inherit the retrieved context as given. A system whose retrieval layer has returned an anachronistic version of a provision can produce an answer that is perfectly faithful to that retrieved text and completely wrong about the legal state on the queried date. In short, these metrics evaluate answer--context alignment; legally adequate evaluation also requires context--law alignment.

The diagnostic criteria proposed in Section~\ref{sec:three_limitations} are intended to be operational rather than merely descriptive. This paper does not report a benchmarked empirical comparison of existing systems; its contribution is the formulation of the diagnostic framework and the derivation of architectural commitments from it. A natural next step is to instantiate the criteria as benchmark tasks. A minimal diagnostic benchmark suite would include six task families:

\begin{enumerate}[leftmargin=2em, itemsep=0.5ex]
    \item \textbf{Structural-context closure:} given a fragment subordinate to a governing clause, recover all ancestors needed for interpretation, and distinguish legally justified expansion from leakage across hierarchical boundaries.
    
    \item \textbf{Descendant completeness under scope queries:} given a query such as ``all items of article $X$'' or ``all sub-items of item $Y$'', return the complete set of represented descendants rather than an approximate Top-$K$ subset.
    
    \item \textbf{Point-in-time and bitemporal correctness:} given a provision and a target date, retrieve the version valid at that date, distinguish valid time from transaction time, and exclude current, consolidated, or temporally adjacent versions that were not operative at the query target.
    
    \item \textbf{Interpretive--statutory coherence:} given a provision whose binding interpretation has shifted without textual amendment, return both the statutory version chain and the interpretive authority chain relevant to the query date, including the operative interpretation and its temporal effect, or flag that the operative state depends on unresolved interpretive authority.
    
    \item \textbf{Provenance-chain reconstruction:} given a current or historical provision, return the ordered chain of legal acts that opened, modified, suspended, interpreted, or closed the queried state, in a form that can be checked against primary sources.
    
    \item \textbf{Claim-level auditability:} when a system synthesizes an answer from retrieved material, associate each material legal claim with the supporting provision, version, interpretation, and grounding event.
\end{enumerate}

Recent work is beginning to formalize adjacent evaluation objectives. Controlling Authority Retrieval, for example, requires systems to retrieve superseding or controlling authority rather than merely relevant documents~\parencite{bacellar2026car}, converging, from a different starting point, on the criterion articulated in this section. The remaining methodological problem is to design metrics that do not reduce institutional traceability to a metadata presence check. The question is not merely whether amendment identifiers are present in the output, but whether the chain of institutional acts is reconstructible and verifiable from the returned state, the attached audit artifact, or a stable follow-up operation. Expert-adjudicated audit exercises may be necessary where automated comparison is too shallow.

A secondary question is how to compare deterministic-by-design systems fairly against probabilistic baselines. The two paradigms optimize for different criteria and may answer different classes of queries correctly. Benchmarks that privilege one paradigm's native output format produce misleading comparisons. Progress requires benchmark designs in which correctness conditions are specified in advance, independently of any reference system: the target date, jurisdiction, structural scope, valid-time interval, provenance chain, and accepted authority set must be part of the benchmark definition rather than inferred from what a baseline happens to retrieve.

\subsection{Cross-Jurisdictional Portability}
\label{subsec:agenda_jurisdictional}

The architectural commitments of Section~\ref{subsec:commitments} are jurisdiction-agnostic in their formulation; their instantiations are not. The trade-off noted in Section~\ref{subsec:tradeoffs} extends naturally into a research agenda because the practical viability of deterministic-by-design retrieval at scale depends on whether the paradigm can generalize across jurisdictions with bounded adaptation cost. The research question is therefore not whether one legal ontology can be reused unchanged everywhere. It is what must remain invariant, what must be localized, and how that localization can be made explicit, auditable, and repeatable.

Three questions structure this front. First, what subset of the ontological and temporal apparatus is genuinely portable across jurisdictions, and what subset requires jurisdiction-specific adaptation? The part--whole structure of legislative drafting is common in modern codified systems, but the relevant units, labels, and legal effects of hierarchy vary across statutes, constitutional texts, regulations, and administrative instruments. Event reification is close to a cross-jurisdictional invariant at the abstract level: legal systems change through acts that create, modify, suspend, interpret, or terminate legal states. But the event subtypes, competent authorities, temporal effects, and conditions of binding force are jurisdiction-specific. Separating invariants from variants is a precondition for any reusable architectural core.

Second, can jurisdictional adaptation be templated? If the shared core is identifiable, the adaptation work for a new jurisdiction reduces to specifying the jurisdiction-specific extensions: drafting conventions, event taxonomies, temporal-effect rules, authority hierarchies, and procedural classes that distinguish binding from non-binding acts. Whether this specification work can be captured in reusable templates, and whether those templates can be filled in by legal experts with limited software expertise, is an open question with direct consequences for adoption.

Third, how do multi-level or hybrid legal corpora fit into a per-jurisdiction architectural model? Hybrid corpora are legal materials whose authority, implementation, or validity depends on more than one legal order, such as EU directives transposed into national law, treaties incorporated into domestic law, or constitutional provisions whose operation depends on supranational or international obligations. These cases are not exotic; they are the daily reality of international and European legal practice. A framework that handles single jurisdictions well but degrades on multi-level corpora leaves substantial territory uncovered.

Progress on this front will likely be collaborative across national and supranational legal-tech communities. Work in single jurisdictions produces instances; comparison across jurisdictions produces theory; and multi-level corpora test whether the theory survives contact with legal systems whose sources of authority are distributed across institutional layers.

\subsection{Human--System Collaboration under Determinism}
\label{subsec:agenda_collaboration}

The three preceding themes concern the substrate and the evaluation of deterministic legal retrieval. A fourth, qualitatively different theme concerns how such systems alter the practice they serve. Probabilistic retrieval in the legal domain is familiar in its failure modes: fluent outputs, occasional fabrications, and human verification as the universal downstream check. Deterministic retrieval has different operational limits: narrower represented coverage, explicit refusals, and flags for unsupported or ambiguous queries. At the same time, it offers a different success signature: machine-verifiable audit trails. These differences change how lawyers interact with the system, where the human remains in the loop, and how confidence is calibrated.

Several open questions follow. First, how does interface design change when a system can distinguish among different kinds of non-answer? ``I do not know'', ``the substrate has no represented state for this query'', ``the query is outside the modeled jurisdiction'', and ``the operative state depends on unresolved interpretation'' are different failures with different legal meanings. Current legal AI interfaces rarely surface these distinctions, partly because probabilistic retrieval cannot reliably make them. Deterministic retrieval, when supported by an explicit substrate and typed failure states, can make these distinctions; the interface-design implications are largely unexplored.

Second, where should humans intervene? In probabilistic retrieval, the canonical answer is often ``everywhere'', because the system may fabricate, retrieve the wrong material, or synthesize plausible content without reliable signals of error. In deterministic retrieval, the division of labor can be more structured. Humans should intervene in the interpretation of retrieved content, in the judgment about whether the retrieved norm applies to the facts at hand, in the resolution of doctrinal ambiguity, and in the quality assurance of the underlying substrate. Formalizing this division of labor as a matter of interface and workflow design is an open research area that intersects AI, law, and human--system interaction.

Third, how do practitioners calibrate trust in a system whose guarantees are architectural rather than statistical? A deterministic retrieval system typically does not express uncertainty as a calibrated probability over semantically similar fragments. It returns a warranted state with its audit trail, reports that no represented state satisfies the query, or flags that the query falls outside the substrate's coverage or rules. This mode is unfamiliar, and it requires a different kind of trust from the one users have developed for probabilistic systems. How that trust is built, tested, and maintained, and what happens when the substrate is incomplete or the ontology is out of date, is a legitimate empirical research question.

A final question cuts across the others. The deterministic-by-design paradigm reallocates responsibility. A probabilistic system that fabricates a citation often leaves the immediate verification burden with the user. A deterministic system that returns wrong content despite a clean audit trail points upstream: to the substrate, the ontology, the source ingestion process, or the maintenance workflow. This transfer of responsibility is the institutional counterpart of the asymmetric error structure discussed in Section~\ref{subsec:tradeoffs}: deterministic systems reduce silent downstream error by moving correctness obligations upstream to substrate design and maintenance. How legal-tech organizations adapt to this reallocation of responsibility --- in liability, workflow, documentation, professional norms, and quality governance --- is a question that will shape deployment as much as any technical consideration.

\paragraph{Closing remark.}
The four themes above are not exhaustive, and the boundaries between them are porous. Work on interpretive time will require benchmark extensions; evaluation methodology will shape cross-jurisdictional comparison; portability will require collaboration practices that connect to the human-system theme; and human-system research will determine whether deterministic guarantees are actually usable in legal practice. Read together, however, the themes sketch a research program in which the three pathologies of Section~\ref{sec:three_limitations} serve as the diagnostic anchor and the four architectural commitments of Section~\ref{subsec:commitments} as the prescriptive one. That program is the productive direction for legal AI research at the retrieval layer over the next several years.

% =============================================================================
\section{Conclusion}
\label{sec:conclusion}

The reliability failures of AI systems in legal contexts have, in public discourse, been attributed mostly to the confabulation tendencies of large language models. The diagnosis is correct but shallow. Behind the fluent fabrications lies a deeper architectural mismatch: the retrieval substrates that feed these models are calibrated for approximate semantic relevance in open-domain corpora, while the legal domain requires validity grounding, temporal correctness, and traceable institutional provenance. This paper has argued that the visible failures of legal AI are surface expressions of this underlying mismatch, and that addressing them requires reframing what retrieval in the legal domain must do.

We developed this reframing in three moves. First, we articulated the \textit{ontological commitment} of legal knowledge: hierarchical and mereological structure, diachronic dynamism under operational closure, and causal traceability of institutional provenance. These commitments are not disciplinary preferences. They are conditions of legal adequacy derived from the classical theory of the legal order and translated into requirements for computational treatment.

Second, we named three pathologies of retrieval that correspond to systematic failures of those commitments: \textit{mereological blindness}, the failure to preserve part--whole structure; \textit{diachronic blindness}, the failure to reconstruct temporal state; and \textit{causal opacity}, the failure to expose the institutional chain behind returned provisions, operative interpretations, or material synthesized claims. The pathologies are architecture-agnostic, and for each we provided operational definitions, failure mechanisms, canonical examples, and detection criteria intended to be reusable as diagnostic tools.

Third, we reviewed the state of the art through the lens of these pathologies rather than through the lens of technologies. The review showed that each pathology is addressed unevenly by identifiable lines of work, but that partial solutions do not compose automatically when retrofitted onto substrates designed for other purposes. Joint satisfaction requires a different architectural posture, which we articulated through four concrete commitments: ontological primacy, event reification, bitemporal correctness, and deterministic interaction protocols. These commitments form a mutually reinforcing package: each contributes to the others, and partial adoption tends to leave residual failure modes at the boundaries.

The overall thesis is that legal AI at the retrieval layer needs more than incremental improvement within the paradigm of probabilistic similarity. To move \textit{beyond probabilistic similarity} as the organizing principle of legal retrieval is not to reject probabilistic methods wholesale; it is to recognize that similarity, whatever its geometry, is an insufficient primitive for a domain whose load-bearing relations concern structure, temporal validity, and institutional provenance. Probabilistic retrieval remains useful for many legal information tasks. It is insufficient, however, as the sole foundation for applications in which legal adequacy is non-negotiable. The alternative we have argued for is deterministic-by-design legal retrieval: architectures grounded in formal ontology, reified events, bitemporal state, and auditable primitives. Such architectures are not replacements for all legal retrieval. They are appropriate substrates for the subset of legal applications in which validity grounding, temporal correctness, and institutional provenance are load-bearing requirements.

Several limitations of this analysis should be acknowledged alongside its contributions. Our treatment concerns \textit{quaestio juris}, not \textit{quaestio facti}; we have not addressed the reconstruction of disputed facts, the evaluation of evidence, or the adjudication of empirical claims. Our scope has centered on statutory and constitutional retrieval, with case law treated as an extension rather than as a first-class domain. Interpretive time is as consequential as statutory time, and we have sketched its shape only enough to indicate where the framework must extend. Finally, the architectural commitments are formulated at a level of abstraction that admits multiple concrete instantiations, and existing instantiations remain uneven across jurisdictions, domains, and evaluation settings.

These limitations are invitations rather than objections. The research agenda laid out in Section~\ref{sec:agenda} identifies four directions --- interpretive time, evaluation methodology, cross-jurisdictional portability, and human--system collaboration --- along which the framework can be tested, extended, and operationalized.

Legal AI is at a juncture. The probabilistic paradigm has produced both impressive demonstrations and well-documented failures. The deterministic-by-design direction argued for here is less visible, but better aligned with what high-stakes legal retrieval demands. The choice is not whether probabilistic methods should disappear from legal AI. It is where the burden of legal correctness should lie: dispersed across countless downstream verifications by users, or built into the retrieval substrate through structure, time, and provenance from the start. This paper has argued for the latter. Whether and how the field moves in that direction is the question the coming years will decide.

% =============================================================================
\printbibliography

\end{document}